\documentclass[lettersize,journal]{IEEEtran}
\usepackage{amsmath,amsfonts}
\usepackage{algorithmic}
\usepackage{algorithm}
\usepackage{amsmath}
\usepackage{amssymb}
\usepackage{amsthm}
\usepackage{array}
\usepackage{hyperref}
\usepackage[caption=false,font=normalsize,labelfont=sf,textfont=sf]{subfig}
\usepackage{textcomp}
\usepackage{stfloats}
\usepackage{url}
\usepackage{verbatim}
\usepackage{graphicx}
\usepackage{cite}
\usepackage{adjustbox}
\usepackage[table]{xcolor}
\usepackage{booktabs}
\usepackage{multirow}
\usepackage{stackrel}
\usepackage{subcaption}
\usepackage{caption}
\usepackage{graphicx}
\usepackage[utf8]{inputenc}

\begin{document}

\title{Diffusion Models, Image Super-Resolution\\And Everything: A Survey}

\author{Brian B. Moser$^{1,2}$, Arundhati S. Shanbhag$^{1,2}$, Federico Raue$^{1}$, Stanislav Frolov$^{1,2}$, Sebastian Palacio$^{1}$, Andreas Dengel$^{1,2}$\\
$^1$ German Research Center for Artificial Intelligence (DFKI), Germany\\
$^2$ Rheinland-Pfälzische Technische Universität Kaiserslautern-Landau, Germany\\
{\tt\small first.second@dfki.de}}


\IEEEpubid{0000--0000/00\$00.00~\copyright~2021 IEEE}

\maketitle

\begin{abstract}
Diffusion Models (DMs) have disrupted the image Super-Resolution (SR) field and further closed the gap between image quality and human perceptual preferences.
They are easy to train and can produce very high-quality samples that exceed the realism of those produced by previous generative methods.
Despite their promising results, they also come with new challenges that need further research: high computational demands, comparability, lack of explainability, color shifts, and more.
Unfortunately, entry into this field is overwhelming because of the abundance of publications.
To address this, we provide a unified recount of the theoretical foundations underlying DMs applied to image SR and offer a detailed analysis that underscores the unique characteristics and methodologies within this domain, distinct from broader existing reviews in the field.
This survey articulates a cohesive understanding of DM principles and explores current research avenues, including alternative input domains, conditioning techniques, guidance mechanisms, corruption spaces, and zero-shot learning approaches. 
By offering a detailed examination of the evolution and current trends in image SR through the lens of DMs, this survey sheds light on the existing challenges and charts potential future directions, aiming to inspire further innovation in this rapidly advancing area.
\end{abstract}


\section{Introduction}
\IEEEPARstart{I}{n} the ever-evolving field of computer vision, the task of image Super-Resolution (SR) – enhancing Low-Resolution (LR) images into High-Resolution (HR) counterparts – stands as a longstanding challenge due to its ill-posed nature: 
Multiple HR images are plausible for any given LR image, differing in aspects such as brightness and color \cite{sun2020learned}.
Its applications span a broad spectrum, from everyday photography to refining satellite \cite{valsesia2021permutation} and medical images \cite{bashir2021comprehensive}.
Despite notable achievements of prior generative SR models, each comes with its own limitations. 
For example, the computational demands of Autoregressive models often outweigh their utility, while Normalizing Flows or Variational Autoencoders struggle to match quality expectations \cite{lee2022autoregressive, esser2021taming, guo2022lar}. 
Although powerful, Generative Adversarial Networks (GANs) need careful regularization and optimization strategies to overcome instability issues \cite{frolov2021adversarial}.

The advent of Diffusion Models (DMs) marks a significant shift in image generation tasks, including SR, challenging the long-standing dominance of Generative Adversarial Networks (GANs) \cite{ho2020denoising, goodfellow2020generative, song2019generative, song2020score}. 
Applications like Dall-E and Stable Diffusion demonstrate that DMs have surpassed GANs in various aspects \cite{dhariwal-diffusion-models-beat-gans, rombach-latent-diffusion-models, ramesh2022hierarchical}. 
Their capability to generate high-quality images from LR inputs has shown immense promise in SR by closely aligning with the qualitative judgments of human evaluators \cite{sr3-google}.
In other words, human raters perceive SR images generated by DMs as more realistic than those produced by other generative models like GANs.

However, as the volume of publications expands, staying updated on the latest developments is becoming more challenging, particularly for those new to the field.
DMs diverge fundamentally from prior generative models and pose new challenges while addressing the limitations of earlier models. 
Identifying coherent trends and potential research directions is challenging despite this rapid expansion. 
This survey aims to demystify DMs, offers a comprehensive overview that bridges foundational concepts with the forefront of image SR and critically analyzes current strengths and weaknesses.

The presented survey builds upon the previous work \textit{Hitchhiker's Guide to Super-Resolution} \cite{10041995}, which gives a broad overview of the image SR field in general.
Similar in spirit is the survey of Li \textit{et al.}, which reviews diffusion models on the more general image restoration tasks like inpainting and dehazing \cite{li2023diffusion}.
Both have overlapping topics, such as the foundations and types of DMs, namely DDPMs \cite{ho2020denoising}, SGMs \cite{song2020score}, and SDEs \cite{song2019generative}.
Moreover, both surveys highlight the introduction of conditioning strategies and zero-shot diffusion as well as show potential research directions.
However, our survey covers all topics related to image SR and is, therefore, more detailed regarding recent developments specifically developed for image SR. 
Moreover, we explain SR-related challenges, like color shifting and cascaded image SR.
We also highlight the relationship of DMs with other generative SR models, namely Variational Autoencoders, GANs, and Flow-based methods.
In addition, we review frequency-based DMs, alternative corruption spaces, and diffusion-based image SR applications.

Concluding with a discussion on emerging trends and their potential for reshaping SR and DM development, this survey sets the stage for future research. 
By offering clarity and direction in the rapidly evolving domain of DMs, we aim to inspire and inform the next wave of research, fostering advancements that continue to push the boundaries of what is possible in image SR with DMs.

The structure of this paper is organized as follows:
\newpage
\noindent
\textbf{\autoref{sec:definitions} - Super-Resolution Basics:} This section provides fundamental definitions and introduces standard datasets, methods, and metrics for assessing image quality commonly utilized in image SR publications.

\noindent
\textbf{\autoref{sec:diffusion-models} - Diffusion Models Basics:} Introduces the principles and various formulations of DMs, including Denoising Diffusion Probabilistic Models (DDPMs), Score-based Generative Models (SGMs), and Stochastic Differential Equations (SDEs). 
This section also explores how DMs relate to other generative models.

\noindent
\textbf{\autoref{sec:improvements} - Improvements for Diffusion Models:} Common practices for enhancing DMs, focusing on efficient sampling techniques and improved likelihood estimation.

\noindent
\textbf{\autoref{sec:diffusionSR} - Diffusion Models for Image SR:} Presents concrete realizations of DMs in SR, explores alternative domains (latent space and wavelet domain), discusses architectural designs and multiple tasks with Null-Space Models, and examines alternative corruption spaces.

\noindent
\textbf{\autoref{sec:aplications} - Domain-Specific Applications:} DM-based SR applications, namely medical imaging, blind face restoration, atmospheric turbulence in face SR, and remote sensing.

\noindent
\textbf{\autoref{sec:discussion} - Discussion and Future Work: } Common problems of DMs for image SR and noteworthy research avenues for DMs specific to image SR.

\noindent
\textbf{\autoref{sec:conclusion} - Conclusion: } 
Summarizes the survey.

\section{Image Super-Resolution}
\label{sec:definitions}
The goal of image Super-Resolution (SR) is to transform one or more Low-Resolution (LR) images into High-Resolution (HR) images. 
The domain can be broadly categorized into two areas \cite{10041995}: Single Image Super-Resolution (SISR) and Multi-Image Super-Resolution (MISR). 
In SISR, a single LR image leads to a single HR image. 
In contrast, MISR methods use multiple LR images to produce one or many HR outputs. 
This section focuses on SISR and explores relevant datasets, established SR models, and techniques to assess image quality.

Given a LR image  $\mathbf{x}  \in \mathbb{R}^{\bar{w} \times \bar{h} \times c}$,  the goal is to generate a HR counterpart $\mathbf{y}  \in \mathbb{R}^{w \times h \times c}$ with $\bar{w} < w$ and $\bar{h} < h$. 
The relationship is represented by a degradation mapping
\begin{equation}
    \label{eq:degradation}
    \mathbf{x} = \mathcal{D} \left( \mathbf{y}; \Theta \right) = ((\mathbf{y}\otimes \boldsymbol{k})\downarrow_s + \boldsymbol{n})_{JPEG_{q}}
\end{equation} 
where $\mathcal{D}$ is a degradation map $\mathcal{D}:\mathbb{R}^{w \times h \times c} \rightarrow \mathbb{R}^{\bar{w} \times \bar{h} \times c}$ and $\Theta$ contains degradation parameters, including aspects like blur $\boldsymbol{k}$, noise $\boldsymbol{n}$, scaling $s$, and compression quality $q$ \cite{liu2022blind}. 
The degradation is typically unknown, posing the main challenge in determining the inverse mapping of $\mathcal{D}$ with parameters $\theta$, usually embodied as SR model.
It leads to an optimization task aimed at minimizing the difference between the predicted SR image $\mathbf{\hat{y}}$ and the original HR image $\mathbf{y}$: 
\begin{equation}
\label{eq:minLoss}
\theta^* = \operatorname{argmin}_{\theta} \mathcal{L} \left( \mathbf{\hat{y}}, \mathbf{y}\right) + \lambda \phi(\theta),
\end{equation}
where $\mathcal{L}$ represents the loss between the predicted SR image and the actual HR image. 
Here, $\lambda$ is a balancing parameter, while $\phi(\theta)$ is introduced as a regularization term.

The inherent complexity arises from the ill-posed nature of predicting $\theta$, as several SR images can be valid for any given LR image, i.e., they can have similar loss values compared to the ground-truth image but are subjectively perceived differently due to many aspects like brightness and coloring \cite{anwar2020densely,sun2020learned,sr3-google}. 
Traditional regression techniques, like standard CNNs, are often adequate for lower magnifications but struggle to replicate high-frequency details required at higher magnifications (e.g., $s > 4$).
To address this, SR models must hallucinate realistic details beyond interpolation, which typically falls under the umbrella topic of generative models, where diffusion models are now at the forefront.

\subsection{Datasets}
\label{sec:datasets}
Several datasets offer a variety of images, resolutions, and content types.
Typically, these datasets consist of LR and HR image pairs.
However, some datasets contain only HR images, with LR images created by bicubic downsampling with anti-aliasing - a default setting for \textit{imresize} in MATLAB \cite{MATLAB:2017b}. 
One famous general SR train set is the Diverse 2K resolution (DIV2K) dataset \cite{div2k}, which includes various realistic images at different resolutions designed specifically for image SR.
Classical test datasets for SR models trained on DIV2K are Set5 \cite{bevilacqua2012low}, Set14 \cite{zeyde2010single}, BSDS100 \cite{martin2001database}, Urban100 \cite{huang2015single} and Manga109 \cite{matsui2017sketch} that cover a variety of scenes and images contents like buildings and manga paintings.
Flickr2K \cite{flickr2k} and Flickr-Faces-HQ (FFHQ) \cite{ffhq} offer diverse sets of human-centric and scene-centric images from Flickr, respectively.
While FFHQ is commonly employed for training models for face SR tasks, Flickr2K is usually used as a train data extension in combination with DIV2K.
Another dataset for face SR is CelebA-HQ \cite{celeba-hq}, which provides high-quality celebrity images and is typically used to evaluate FFHQ-trained SR models.
For broader applications in CV, datasets like ImageNet \cite{imagenet} and Visual Object Classes (VOC2012) \cite{voc-2012} are favored. 
ImageNet offers an extensive range of images that help train models on various object classes, whereas VOC2012 is vital for object detection and segmentation. 
Both are valuable for multi-task learning involving SR.
More datasets can be found in the \textit{Hitchhiker's Guide to Super-Resolution} \cite{10041995}.
%

\subsection{SR Models}
The primary objective is to design a SR model $\mathcal{M}: \mathbb{R}^{\bar{w} \times \bar{h} \times c} \rightarrow \mathbb{R}^{w \times h \times c}$, such that it inverses \autoref{eq:degradation}: 
\begin{equation}
    \mathbf{\hat{y}} = \mathcal{M} \left( \mathbf{x}; \theta \right),
\end{equation}
where $\mathbf{\hat{y}}$ is the predicted HR approximation of the LR image $\mathbf{x}$ and $\theta$ the parameters of $\mathcal{M}$. 
The parameters $\theta$ are optimized using \autoref{eq:minLoss}, i.e., minimizing the loss function $\mathcal{L}$ between the estimation $\mathbf{\hat{y}}$ and the ground-truth HR image $\mathbf{y}$. 
The following section focuses on standard methods for designing an SR model, especially deep learning methods before we examine how diffusion models fulfill this role in detail. 

\label{sec:existing-work}
\textbf{Traditional Methods:}
Traditional methods for image SR define a range of methodologies, such as statistical \cite{stat-based-1}, edge-based \cite{edge-based-1}, \cite{edge-based-2} patch-based \cite{patch-based-1}, \cite{patch-based-2}, prediction-based \cite{pred-based-1}, \cite{pred-based-2} and sparse representation techniques \cite{sparse-based-1}. 
They fundamentally rely on image statistics and the information inherent in existing pixels to generate HR images.
Despite their utility, a noteworthy drawback of these methods is the potential introduction of noise, blur, and visual artifacts \cite{10041995}.

\textbf{Regression-based Deep Learning:}
Image SR significantly evolved with advancements in deep learning and computational power. 
Typically, they employ a Convolutional Neural Network (CNN) for end-to-end mapping from LR to HR. 
Initial models, such as SRCNN \cite{srcnn}, FSRCNN \cite{fsr-cnn}, and ESPCNN \cite{espcnn}, utilized simple CNNs of diverse depth and feature maps sizes.
Later models adapted concepts from the broader CV domain into SR models, e.g., ResNet led to SRResNet, where residual information was propagated to successive network layers \cite{srresnet}.
Likewise, DenseNet\cite{densenet} was adapted with SRDenseNet \cite{srdensenet}. 
They employ dense blocks, where each layer receives additionally the features generated in all preceding layers.
Recursive CNNs that recursively use the same module to learn representations were also inspired by other CV methods for regression-based SR methods in DRCN \cite{drcn}, DRRN \cite{drrn}, and CARN \cite{carn}. 
More recently, attention mechanisms have been incorporated to focus on regions of interest in images, predominantly via the channel and spatial attention mechanisms \cite{10041995, liang2021swinir, chen2023activating, hsu2024drct}.  
All those methods have in common that they are regression-based.
Commonly used loss functions are the L1 and L2 losses. 
As mentioned, they often produce satisfying results for lower magnifications but struggle to replicate the high-frequency details required at higher magnifications (e.g., $s>4$).
These limitations arise because these models primarily learn an averaged mapping (due to L1 and L2 losses) from LR to HR images, which tends to produce overly smooth textures lacking detail, especially noticeable in larger upscaling factors \cite{10041995}.
To address this, SR models must hallucinate realistic details beyond simple interpolation, a challenge typically tackled by generative models.
%

\textbf{Generative Adversarial Networks (GANs):}
One of the most prominent generative models is the Generative Adversarial Network (GAN). 
It uses two CNNs: A generator G and a discriminator D, which are trained simultaneously. 
The generator aims to produce HR samples that are as close to the original as to fool the discriminator, which tries to distinguish between generated and real samples. 
This framework, e.g., in SRGAN \cite{srgan} or ESRGAN \cite{esrgan}, is optimized using a combination of adversarial loss and content loss to produce less-smoothed images. 
The resultant images of state-of-the-art GANs are sharper and more detailed. 
Due to their capability to generate high-quality and diverse images, they have received much attention lately.
However, they are susceptible to mode collapse, have a sizeable computational footprint, sometimes fail to converge, and suffer from stabilization issues \cite{frolov2021adversarial}. 

\textbf{Flow-based Methods:}
Flow-based methods employ optical flow algorithms to generate SR images \cite{srflow}. 
They were introduced in an attempt to counter the ill-posed nature of image SR by learning the conditional distribution of plausible HR images given a LR input. 
They introduce a conditional normalized flow architecture that aligns LR and HR images by calculating the displacement field between them and then uses this information to recover SR images. 
They employ a fully invertible encoder capable of mapping any input HR image to the latent flow space and ensuring exact reconstruction. 
This framework enables the SR model to learn rich distributions using exact log-likelihood-based training \cite{srflow}. 
This facilitates flow-based methods to circumvent training instability but incurs a substantial computational cost.

\subsection{Image Quality Assessment (IQA)}
\label{sec:iqa}
Image quality is a multifaceted concept that addresses properties like sharpness, contrast, and absence of noise. 
Hence, a fair evaluation of SR models based on produced image quality forms a non-trivial task.
This section presents the essential methods, especially for diffusion models, to assess image quality in the context of image SR, which fall under the umbrella term Image Quality Assessment (IQA) \footnote{More SR-related IQA methods can be found in Moser \textit{et al.} \cite{10041995}.}.
At its core, IQA refers to any metric that resembles the perceptual evaluations from human observers, specifically, the level of realism perceived in an image after the application of SR techniques. 
During this section, we will use the following notation: $N_\mathbf{x} = w \cdot h \cdot c$, which defines the number of pixels of an image $\mathbf{x} \in \mathbb{R}^{w \times h \times c}$ and $\Omega_\mathbf{x} = \{ \left( i,j,k \right) \in \mathbb{N}_1^3 | i \leq h, j \leq w, k \leq c \}$ that defines the set of all valid positions in $\mathbf{x}$.

\textbf{Peak Signal-to-Noise Ratio (PSNR):}
The Peak Signal-to-Noise Ratio (PSNR) is one of the most widely used techniques to evaluate SISR reconstruction quality. 
It represents the ratio between the maximum pixel value L and the Mean Squared Error (MSE) between the SR image $\mathbf{\hat{y}}$ and the HR image $\mathbf{y}$.  
\begin{equation}
    \text{PSNR}  \left( \mathbf{y}, \mathbf{\hat{y}} \right) = 10 \cdot \log_{10} \left( \frac{L^2}{\frac{1}{N} \sum _{i=1}^{N} \left[ \mathbf{y} - \mathbf{\hat{y}} \right]^2} \right)
    \label{eq:psnr}
\end{equation}
Despite being one of the most popular IQA methods, it does not accurately match human perception \cite{sr3-google}. 
It focuses on pixel differences, which can often be inconsistent with the subjectively perceived quality: the slightest shift in pixels can result in worse PSNR values while not affecting human perceptual quality.
Due to its pixel-level calculation, models trained with correlated pixel-based loss tend to achieve high PSNR values \cite{10041995}, whereas generative models tend to produce lower PSNR values \cite{sr3-google}.

\textbf{Structural Similarity Index (SSIM):}
The SSIM, like the PSNR, is a popular evaluation method that focuses on the differences in structural features between images. 
It independently captures the structural similarity by comparing luminance, contrast, and structures. SSIM estimates for an image $\mathbf{y}$ the luminance $\mu_\mathbf{y}$ as the mean of the intensity, while it is estimating contrast $\sigma_\mathbf{y}$ as its standard deviation:
\begin{equation}
    \mu_\mathbf{y} = \frac{1}{N_\mathbf{y}} \sum_{p \in \Omega_\mathbf{y}} \mathbf{y}_p,
\end{equation}
\begin{equation}
    \sigma_\mathbf{y} = \frac{1}{N_\mathbf{y} - 1} \sum_{p \in \Omega_\mathbf{y}} \left[ \mathbf{y}_p - \mu_\mathbf{y}\right]^2
\end{equation}
To capture the similarity between the computed entities, the authors introduced a comparison function $S$:
\begin{equation}
    \label{eq:simfun}
    S \left( x, y, c\right) = \frac{2 \cdot x \cdot y + c}{x^2 + y^2 + c}\,,
\end{equation}
where $x$ and $y$ are the scalar variables being compared, and $c = \left(k \cdot L \right)^2$, $0 < k \ll 1$ is a constant for numerical stability.
For a HR image $\mathbf{y}$ and its approximation $\mathbf{\hat{y}}$, the luminance ($\mathcal{C}_l$) and contrast ($\mathcal{C}_c$) comparisons are computed using $\mathcal{C}_l \left( \mathbf{y}, \mathbf{\hat{y}} \right) = 
S \left( \mu_\mathbf{y}, \mu_\mathbf{\hat{y}}, c_1\right)$ and $ \mathcal{C}_c \left( \mathbf{y}, \mathbf{\hat{y}} \right) = S \left( \sigma_\mathbf{y}, \sigma_\mathbf{\hat{y}}, c_2 \right)$, where $c_1, c_2 > 0$.
The empirical covariance
\begin{equation}
    \sigma_{\mathbf{y}, \mathbf{\hat{y}}} = \frac{1}{ N_\mathbf{y} - 1} \sum_{p \in \Omega_\mathbf{y}} \left( \mathbf{y}_p- \mu_\mathbf{y} \right) \cdot \left( \mathbf{\hat{y}}_p - \mu_\mathbf{\hat{y}} \right),
\end{equation}
defines the structure comparison ($\mathcal{C}_s$), which is the correlation coefficient between $\mathbf{y}$ and $\mathbf{\hat{y}}$:
\begin{equation}
    \mathcal{C}_s \left( \mathbf{y}, \mathbf{\hat{y}} \right) = \frac{\sigma_{\mathbf{y}, \mathbf{\hat{y}}} + c_3}{\sigma_\mathbf{y} \cdot \sigma_\mathbf{\hat{y}} + c_3},
\end{equation}
where $c_3 > 0$. Finally, the SSIM is defined as:
\begin{equation}
    \label{eq:ssim}
    \text{SSIM} \left( \mathbf{y}, \mathbf{\hat{y}}\right) = \left[ \mathcal{C}_l \left( \mathbf{y}, \mathbf{\hat{y}} \right)\right]^\alpha \cdot \left[ \mathcal{C}_c \left( \mathbf{y}, \mathbf{\hat{y}} \right)\right]^\beta \cdot\left[ \mathcal{C}_s \left( \mathbf{y}, \mathbf{\hat{y}} \right)\right]^\gamma
\end{equation}
where $\alpha > 0, \beta > 0$, and $\gamma > 0$ are parameters that can be adjusted to tune the relative importance of the components.

\textbf{Mean Opinion Score (MOS):}
The MOS is a subjective measure that leverages human perceptual quality for the evaluation of the generated SR images. Human viewers are shown SR images and asked to rate them with quality scores that are then mapped to numerical values and later averaged. Typically, these range from 1 (bad) to 5 (good) but may vary \cite{sr3-google}. While this method is a direct evaluation of human perception, it is more time-consuming and cumbersome to conduct compared to objective metrics. Moreover, due to the highly subjective nature of this metric, it is susceptible to bias.  

\textbf{Consistency:} 
Consistency measures the degree of stability of non-deterministic SR methods, such as generative models like GANs or DMs. 
Like flow-based methods, generative approaches are intentionally designed to generate a spectrum of plausible outputs for the same input. 
However, low consistency is not desirable.
Minor variations lessen the influence of a relatively consistent method in the input.
Nevertheless, consistency can vary depending on the requirements.
One commonly employed metric to quantify consistency is the Mean Squared Error.

\textbf{Learned Perceptual Image Patch Similarity (LPIPS):}
Contrary to the pixel-based evaluation of PSNR and SSIM, the Learned Perceptual Image Patch Similarity (LPIPS) utilizes a pre-trained CNN $\varphi$, e.g., VGG \cite{simonyan2014very} or AlexNet \cite{krizhevsky2017imagenet}, and generates $L$ feature maps from the SR and HR image, and subsequently calculates the similarity between them.
Given $h_l$ and $w_l$ as the height and width of the $l$-th feature map respectively, and a scaling vector $\alpha_l \in \mathbb{R}^{C_l}$, the LPIPS metric is formulated as follows:
\begin{equation}
\text{LPIPS} \left( \mathbf{y}, \mathbf{\hat{y}}\right) = \sum^L_{l=1} \sum_{p }\frac{ \left\lVert \alpha_l \odot \left( \varphi^l \left( \mathbf{\hat{y}} \right) - \varphi^l \left(  \mathbf{y} \right)\right)_p \right\rVert^2_2}{h_l \cdot w_l}
\end{equation}
LPIPS operates by projecting images into a perceptual feature space through $\varphi$ and evaluating the difference between corresponding patches in SR and HR images, scaled by $\alpha_l$. 
This methodology allows for a more human-centric evaluation, given that it is better aligned with human perception than traditional metrics such as PSNR and SSIM \cite{10041995}.


\textbf{No-Reference Metrics:}
All IQA metrics discussed so far require a reference (ground-truth) image.
However, there are cases where no reference images are available, e.g., in unsupervised settings.
Fortunately, we can assess an image by measuring the distance of statistical features from those obtained from a collection of high-quality images of a similar domain, i.e., natural images.
This can be opinion- and distortion-aware like BRISQUE \cite{mittal2012no} or opinion- and distortion-unaware like NIQE \cite{mittal2012making}.
Another intriguing way to assess no-reference image quality is to exploit the visual-language pre-trained CLIP model \cite{radford2021learning}.
One example is CLIP-IQA, which calculates the cosine similarity of the encoded image with two prompts of opposing meaning, i.e., "good photo" and "bad photo" \cite{wang2023exploring}. 
The resulting relative similarity metric for one or the other prompt determines the image quality. 
CLIP-IQA shows results comparable to those of BRISQUE without the hand-crafted features and surpasses other no-reference IQA methods like NIQE. 
Another way to exploit deep learning models is to train them to predict subjective scores using IQA datasets like TID2013 \cite{ponomarenko2015image}.
Examples are DeepQA \cite{kim2017deep}, NIMA \cite{talebi2018nima}, or MUSIQ \cite{ke2021musiq}. 
Others can be found in the learning-based perceptual quality section of the \textit{Hitchhiker's Guide to Super-Resolution} \cite{10041995}.

\section{Diffusion Models Basics}
\label{sec:diffusion-models} 

\begin{figure}[!t]
    \begin{center}
        \includegraphics[width=.49\textwidth]{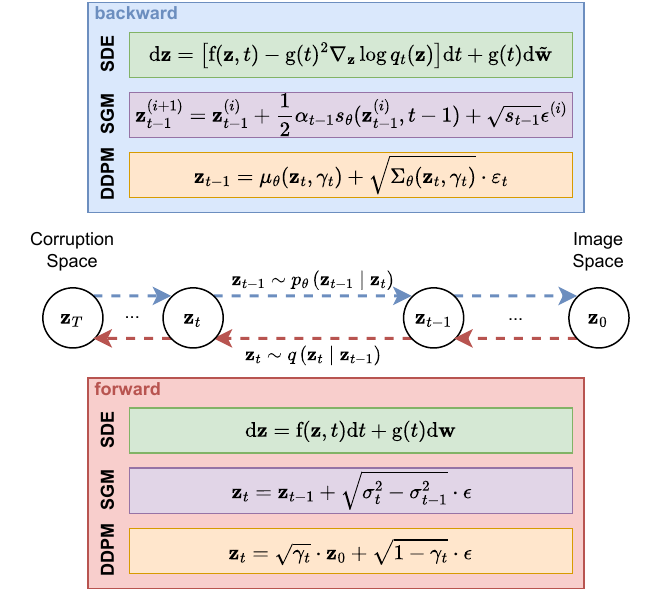}
        
        \caption{\label{fig:dms}
        Principle of DMs. 
        The forward diffusion adds noise iteratively (red), which translates an image from the image space to the corruption space. 
        The backward diffusion, the iterative refinement process, reverts the process (blue) back to the image space. 
        Shown are three different implementations of DMs, namely Denoising Diffusion Probabilistic Models (DDPMs), Score-based Generative Models (SGMs), and Stochastic Differential Equations (SDEs) with their respect formulation of the forward and backward diffusion.
        }
    \end{center}
\end{figure}

Diffusion Models (DMs) have profoundly impacted the realm of generative AI, and many approaches that fall under the umbrella term DM have emerged. 
What sets DMs apart from earlier generative models is their execution over iterative time steps, both forward and backward in time and denoted by $t$, as depicted in \autoref{fig:dms}. 
The forward and backward diffusion processes are distinguished by:

\noindent
\textbf{Forward} $q$ - degrade input data using noise iteratively, forward in time (i.e., $t$ increases).

\noindent
\textbf{Backward} $p$ - denoise the degraded data, thereby reversing the noise iteratively, backward in time (i.e., $t$ decreases).

The time step $t$ increases during forward diffusion, whereas it propagates towards 0 during backward diffusion.
Let $\mathcal{D} = \{\mathbf{x}_i, \mathbf{y}_i\}^N_{i=1}$ be a dataset of LR-HR image-pairs. 
For each time step $t$, the random variable $\mathbf{z}_t$ describes the current state, a state between the image and corruption space.
In literature, there is no clear distinction between $\mathbf{z}_t$ in the forward and $\mathbf{z}_t$ in the backward diffusion.
During forward diffusion, we assume $\mathbf{z}_t \sim q \left( \mathbf{z}_t \mid \mathbf{z}_{t-1} \right)$.
Conversely, in the backward diffusion, we assume $\mathbf{z}_{t-1} \sim p \left( \mathbf{z}_{t-1} \mid \mathbf{z}_t \right)$. 
We will denote $T$ with $0 < t \leq T$ as the maximal time step for finite cases.
The initial data distribution ($t=0$) is represented by $\mathbf{z}_0 \sim q \left( \mathbf{x}\right)$, which is then slowly injected with noise (additive). 
Vice versa, DMs remove noise therein by running a parameterized model $p_\theta \left( \mathbf{z}_{t-1} \mid \mathbf{z}_t \right)$ in the reverse time direction that approximates the ideal (but unattainable) denoised distribution $p \left( \mathbf{z}_{t-1} \mid \mathbf{z}_t \right)$. 

The explicit implementation of the forward diffusion $q$ and backward diffusion $p$, approximated by $p_\theta$, is defined by the specific DM in use.
There are three types:
Two discrete forms, namely Denoising Diffusion Probabilistic Models (DDPMs) and Score-Based Generative Models (SGMs), and the continuous form by Stochastic Differential Equations (SDEs) \cite{yang2023diffusion}. 
Each of these types will be discussed next are comprehensively shown in \autoref{fig:dms}.

\subsection{Denoising Diffusion Probabilistic Models (DDPMs)}

\label{sec:ddpms}
Denoising Diffusion Probabilistic Models (DDPMs) \cite{ho2020denoising} use two Markov chains to enact the forward and backward diffusion across a finite amount of discrete time steps.

\textbf{Forward Diffusion:} It transforms the data distribution into a prior distribution, typically designed manually (e.g., Gaussian), given by:
\begin{equation}
    q(\mathbf{z}_{t} \mid \mathbf{z}_{t-1}) = \mathcal{N}(\mathbf{z}_{t} \mid \sqrt{1-\alpha_t}\, \mathbf{z}_{t-1}, \alpha_t \mathbf{I} ),
\end{equation}
where the hyper-parameters $0 < \alpha_{1:T} < 1$ represent the variance of noise incorporated at each time step. 
While the Gaussian kernel is commonly adopted, alternative kernel types can also be employed. 
This formulation can be condensed to a single-step calculation, as shown by:
\begin{equation}
    q(\mathbf{z}_t \mid \mathbf{z}_0) = \mathcal{N}(\mathbf{z}_t \mid \sqrt{\gamma_t}\, \mathbf{z}_0, (1-\gamma_t) \mathbf{I}),
\end{equation}
where $\gamma_t = \prod_{i=1}^t (1-\alpha_i)$ \cite{sohl2015deep}. 
Consequently, $\mathbf{z}_t$ can be directly sampled regardless of what ought to happen on previous time steps by 
\begin{equation}
    \label{eq_sr3yt}
    \mathbf{z}_t = \sqrt{\gamma_t} \cdot \mathbf{z}_0  + \sqrt{1-\gamma_t} \cdot \epsilon, \quad \epsilon \sim \mathcal{N} \left( \mathbf{0}, \mathbf{I} \right). 
\end{equation}

\textbf{Backward Diffusion:} The goal is to directly learn the inverse of the forward diffusion and generate a distribution that resembles the prior $\mathbf{z}_0$, usually the HR image in SR.
In practice, we use a CNN to learn a parameterized form of $p$.
Since the forward process approximates $q(\mathbf{z}_{T}) \approx \mathcal{N} \left( \mathbf{0}, \mathbf{I} \right)$, the formulation of the learnable transition kernel becomes:
\begin{equation}
    p_\theta \left( \mathbf{z}_{t-1} \mid \mathbf{z}_t \right) = \mathcal{N} \left( \mathbf{z}_{t-1} \mid \mu_{\theta}(\mathbf{z}_{t}, \gamma_t), \Sigma_\theta(\mathbf{z}_{t}, \gamma_t) \right),
\end{equation}
where $\mu_{\theta}$ and $\Sigma_\theta$ are learnable. 
Similarly, the conditional formulation $p_\theta \left( \mathbf{z}_{t-1} \mid \mathbf{z}_t, \mathbf{x} \right)$ conditioned on $\mathbf{x}$ (e.g., a LR image) is using $\mu_{\theta}(\mathbf{z}_{t}, \mathbf{x}, \gamma_t)$ and $\Sigma_\theta(\mathbf{z}_{t}, \mathbf{x}, \gamma_t)$ instead.

\textbf{Optimization:} To guide the backward diffusion in learning the forward process, we minimize the Kullback-Leibler (KL) divergence of the joint distribution of the forward and reverse sequences 
\begin{align}
    p_\theta \left( \mathbf{z}_{0}, ..., \mathbf{z}_{T}\right) &= p \left(\mathbf{z}_{T}\right)\prod^T_{t=1}p_\theta \left( \mathbf{z}_{t-1} \mid \mathbf{z}_t \right),\text{ and} \\
    q \left( \mathbf{z}_{0}, ..., \mathbf{z}_{T}\right) &= q \left(\mathbf{z}_{0}\right)\prod^T_{t=1}q \left( \mathbf{z}_{t} \mid \mathbf{z}_{t-1} \right),
\end{align}
which leads to minimizing 
\begin{align}
\label{eq:vlb_ddpm}
    &\text{KL} ( q \left( \mathbf{z}_{0}, ..., \mathbf{z}_{T}\right) \| p_\theta \left( \mathbf{z}_{0}, ..., \mathbf{z}_{T}\right)) \\
    &=-\mathbb{E}_{q \left( \mathbf{z}_{0}, ..., \mathbf{z}_{T}\right)} \left[ \log p_\theta \left( \mathbf{z}_{0}, ..., \mathbf{z}_{T}\right) \right] + c \nonumber\\
    &\stackrel{(i)}{=} \mathbb{E}_{q \left( \mathbf{z}_{0}, ..., \mathbf{z}_{T}\right)} \left[ - \log p \left(\mathbf{z}_{T}\right) - \sum^T_{t=1} \log \frac{p_\theta \left( \mathbf{z}_{t-1} \mid \mathbf{z}_t \right)}{q \left( \mathbf{z}_{t} \mid \mathbf{z}_{t-1} \right)}\right] + c \nonumber\\
    &\stackrel{(ii)}{\geq} \mathbb{E} \left[ - \log p_\theta \left( \mathbf{z}_{0}\right)\right] + c, \nonumber
\end{align}
where $(i)$ is possible because both terms are products of distributions and $(ii)$ is the product of Jensen's inequality. 
The constant $c$ is unaffected and, therefore, irrelevant in optimizing $\theta$.
Note that \autoref{eq:vlb_ddpm} without $c$ is the Variational Lower Bound (VLB) of the log-likelihood of the data $\mathbf{z}_0$, which is commonly maximized by DDPMs. 

\subsection{Score-based Generative Models (SGMs)}
Score-based Generative Models (SGMs), much like DDPMs, utilize discrete diffusion processes but employ an alternative mathematical foundation.
Instead of using probability density function $p (\mathbf{z})$ directly, Song \textit{et al.} \cite{song2020score} propose to work with its (Stein) score function, which is defined as the gradient of the log probability density $\nabla_\mathbf{z} \log p(\mathbf{z})$.
Mathematically, the score function preserves all information about the density function, but computationally, it is easier to work with.
Furthermore, the decoupling of model training from the sampling procedure grants greater flexibility in defining sampling methods and training objectives.

\textbf{Forward Diffusion:} Let $0 < \sigma_1 < ... < \sigma_T$ be a finite sequence of noise levels. 
Like DDPMs, the forward diffusion, typically assigned to a Gaussian noise distribution, is 
\begin{equation}
    \label{eq:sgm_q}
    q(\mathbf{z}_t \mid \mathbf{z}_0) = \mathcal{N}(\mathbf{z}_t \mid \mathbf{z}_0, \sigma^2_t \mathbf{I}).
\end{equation}
This equation results in a sequence of noisy data densities $q(\mathbf{z}_1), ..., q(\mathbf{z}_T)$ with $q(\mathbf{z}_t) = \int q(\mathbf{z}_t)q(\mathbf{z}_0)\text{d}\mathbf{z}_0$.
Consequently, the intermediate step $\mathbf{z}_t = \mathbf{z}_0  + \sigma_t \cdot \epsilon$ with $\epsilon \sim \mathcal{N} \left( \mathbf{0}, \mathbf{I} \right)$ can be sampled agnostic from previous time steps in a single step.

\textbf{Backward Diffusion:} To revert the noise during the backward diffusion, we need to approximate $\nabla_{\mathbf{z}_t} \log{q(\mathbf{z}_t)}$ and choose a method for estimating the intermediate states $\mathbf{z}_t$ from that approximation.
For the gradient approximation at each time step $t$, we use a trained predictor, denoted as $s_\theta$ and called Noise-Conditional Score Network (NCSN), such that $s_\theta(\mathbf{z}_t, t) \approx \nabla_{\mathbf{z}_t} \log{q(\mathbf{z}_t)}$ \cite{song2020score}. 

The training of the NCSN will be covered in the next section; for now, we focus on the sampling process using NCSN. 
Sampling with NCSN involves generating the intermediate states $\mathbf{z}_t$ through an iterative approach, using $s_\theta(\mathbf{z}_t, t)$.
Note that this iterative process is different from the iterations done during the diffusion as it addresses solely the generation of $\mathbf{z}_t$.
This is a key difference to DDPMs as $\mathbf{z}_t$ needs to be sampled iteratively, whereas DDPMs directly predict $\mathbf{z}_t$ from $\mathbf{z}_{t+1}$.

There are various ways to perform this iterative generation, but we will concentrate on a specific method known as Annealed Langevin Dynamics (ALD), introduced by Song et al. \cite{song2019generative}.
Let $N$ be the number of estimation iterations for $\mathbf{z}_t$ at time step $t$ and $\alpha_t > 0$ the corresponding step size, which determines how much the estimation moves from one estimate $\mathbf{z}_{t-1}^{(i)}$ towards $\mathbf{z}_{t-1}^{(i+1)}$. 
The initial state is $\mathbf{z}_T^{(N)} \sim \mathcal{N} \left( \mathbf{0}, \mathbf{I} \right)$.
For each $0 < t \leq T$, we initialize $\mathbf{z}_{t-1}^{(0)} = \mathbf{z}_t^{(N)} \approx \mathbf{z}_t$, which is the latest estimation of the previous intermediate state. 
In order to get $\mathbf{z}_{t-1}^{(N)} \approx \mathbf{z}_{t-1}$ iteratively, ALD uses the following update rules for $i = 0, ..., N-1$:
\begin{align}
    \epsilon^{(i)} &\leftarrow \mathcal{N} \left( \mathbf{0}, \mathbf{I} \right) \\
    \mathbf{z}_{t-1}^{(i+1)} &\leftarrow \mathbf{z}_{t-1}^{(i)} + \frac{1}{2}\alpha_{t-1}s_\theta(\mathbf{z}_{t-1}^{(i)}, t-1) + \sqrt{s_{t-1}}\epsilon^{(i)}
\end{align}
This update rule guarantees that $\mathbf{z}_0^{(N)}$ converges to $q(\mathbf{z}_0)$ for $\alpha_t \rightarrow 0$ and $N \rightarrow \infty$ \cite{parisi1981correlation}. 

Similar to DDPMs, we can turn SGMs into conditional SGMs by integrating the condition $\mathbf{x}$, e.g., a LR image, into $s_\theta(\mathbf{z}_{t}, \mathbf{x}, t) \approx \nabla_{\mathbf{z}_t} \log{q(\mathbf{z}_t | \mathbf{x})}$.

\textbf{Optimization:} Without specifically formulating the backward diffusion, we can train a NCSN such that $s_\theta(\mathbf{z}_t, t) \approx \nabla_{\mathbf{z}_t} \log{q(\mathbf{z}_t)}$.
Estimating the score can be done by using the denoising score matching method \cite{vincent2011connection}: 
\begin{align}
    \label{eq:sgm_lastOpt}
    &\mathop{\mathbb{E}}_{\substack{t\sim\mathcal{U}(1,T) \\ \mathbf{z}_0 \sim q( \mathbf{z}_0) \\ \mathbf{z}_t \sim q( \mathbf{z}_t \mid \mathbf{z}_0)}} \left[ \lambda(t)\sigma^2_t \| \nabla_{\mathbf{z}_t} \log{q(\mathbf{z}_t)} - s_\theta(\mathbf{z}_t, t)\|^2\right] \\
    \stackrel{(i)}{=}&\mathop{\mathbb{E}}_{\substack{t\sim\mathcal{U}(1,T) \nonumber \\ \mathbf{z}_0 \sim q( \mathbf{z}_0) \\ \mathbf{z}_t \sim q( \mathbf{z}_t \mid \mathbf{z}_0)}} \left[ \lambda(t)\sigma^2_t \| \nabla_{\mathbf{z}_t} \log{q(\mathbf{z}_t | \mathbf{z}_0)} - s_\theta(\mathbf{z}_t, t)\|^2\right] + c\\
    \stackrel{(ii)}{=}&\mathop{\mathbb{E}}_{\substack{t\sim\mathcal{U}(1,T) \nonumber\\ \mathbf{z}_0 \sim q( \mathbf{z}_0) \\ \mathbf{z}_t \sim q( \mathbf{z}_t \mid \mathbf{z}_0)}} \left[ \lambda(t) \| - \frac{\mathbf{z}_t - \mathbf{z}_0}{\sigma_t} - \sigma_t s_\theta(\mathbf{z}_t, t)\|^2\right] + c \\
    \stackrel{(iii)}{=}&\mathop{\mathbb{E}}_{\substack{t\sim\mathcal{U}(1,T) \nonumber\\ \mathbf{z}_0 \sim q( \mathbf{z}_0) \\ \epsilon \sim \mathcal{N} \left( \mathbf{0}, \mathbf{I} \right)}} \left[ \lambda(t) \| \epsilon + \sigma_t s_\theta(\mathbf{z}_t, t)\|^2\right] + c
\end{align}
where $\lambda(t) > 0$ is a weighting function, $\sigma_t$ the noise level added at time step $t$, $(i)$ derived by Vincent \textit{et al.} \cite{vincent2011connection}, $(ii)$ from \autoref{eq:sgm_q}, $(iii)$ from $\mathbf{z}_t = \mathbf{z}_0 + \sigma_t\epsilon$ and with $c$ again a constant unaffected in the optimization of $\theta$. 
Note that there are other ways to estimate the score, e.g., based on score matching \cite{hyvarinen2005estimation} or sliced score matching \cite{song2020sliced}.

\subsection{Stochastic Differential Equations (SDEs)}
So far, we have discussed DMs that deal with finite time steps. 
A generalization to infinite continuous time steps is made by formulating these as solutions to Stochastic Differential Equations (SDEs), also known as Score SDEs \cite{song2019generative}.
In fact, we can view SGMs and DDPMs as discretizations of a continuous-time SDE.
SDEs are not entirely bound to DMs, as they are a mathematical concept describing stochastic processes.
As such, they fit perfectly to describe the processes we want to simulate in DMs.
Like previously, data is perturbed in a general diffusion process but generalized to an infinite number of noise scales.

\textbf{Forward Diffusion:} We can represent the forward diffusion by the following SDE:
\begin{equation}
    \text{d}\mathbf{z} = \text{f}(\mathbf{z}, t)\text{d}t + \text{g}(t)\text{d}\mathbf{w},
\end{equation}
where $\text{f}$ and $\text{g}$ are the drift and diffusion functions, respectively, and $\mathbf{w}$ is the standard Wiener process (also known as Brownian motion). 
This generalized formulation allows uniform representation of both DDPMs and SGMs.
The SDE for DDPMs is given by:
\begin{equation}
    \text{d}\mathbf{z} = - \frac{1}{2} \alpha(t)\mathbf{z}\text{d}t + \sqrt{\alpha(t)}\text{d}\mathbf{w},
\end{equation}
with $\alpha(\frac{t}{T}) = T\alpha_t$ for $T \rightarrow\infty$.
For SGMs, the SDE is 
\begin{equation}
    \text{d}\mathbf{z} = \sqrt{\frac{\text{d}\left[ \sigma(t)^2\right]}{\text{d}t}}\text{d}\mathbf{w}, 
\end{equation}
with $\sigma(\frac{t}{T}) = \sigma_t$ for $T \rightarrow\infty$. 
From now on, we denote with $q_t(\mathbf{z})$ the distribution of $\mathbf{z}_t$ in the diffusion process.

\textbf{Backward Diffusion:} The reverse-time SDE is formulated by Anderson \textit{et al.} \cite{anderson1982reverse} as:
\begin{equation}
    \label{eq_revProcSDE}
    \text{d}\mathbf{z} =  \left[ \text{f}(\mathbf{z}, t) - \text{g}(t)^2\nabla_\mathbf{z} \log q_t(\mathbf{z})\right]\text{d}t + \text{g}(t)\text{d}\tilde{\mathbf{w}},
\end{equation}
where $\tilde{\mathbf{w}}$ is the standard Wiener process when time flows backwards and $\text{d}t$ an infinitesimal negative time step.
Solutions to \autoref{eq_revProcSDE} can be viewed as diffusion processes that gradually convert noise to data.
The existence of a corresponding probability flow Ordinary Differential Equation (ODE), whose trajectories possess the same marginals as the reverse-time SDE, was proven by Song \textit{et al.} \cite{song2020score} and is 
\begin{equation}
    \text{d}\mathbf{z} =  \left[ \text{f}(\mathbf{z}, t) - \frac{1}{2}\text{g}(t)^2\nabla_\mathbf{z} \log q_t(\mathbf{z})\right]\text{d}t.
\end{equation}
Thus, the reverse-time SDE and the probability flow ODE enable sampling from the same data distribution.

\textbf{Optimization:} Similar to the approach in SGMs, we define a score model such that $s_\theta(\mathbf{z}_t, t) \approx \nabla_{\mathbf{z}} \log{q_t(\mathbf{z})}$.
Additionally, we extend \autoref{eq:sgm_lastOpt} to continuous time as follows:
\begin{equation}
    \mathop{\mathbb{E}}_{\substack{t\sim\mathcal{U}(0,T) \\ \mathbf{z}_0 \sim q( \mathbf{z}_0) \\  \mathbf{z}_t \sim q( \mathbf{z}_t \mid  \mathbf{z}_0)}} \left[ \lambda(t) \| s_\theta(\mathbf{z}_t, t) - \nabla_{\mathbf{z}_t} \log{q_t(\mathbf{z}_t \mid \mathbf{z}_0)}\|^2\right],
\end{equation}
where $\lambda(t) > 0$ is a weighting function.

\begin{figure}
    \begin{center}
        \includegraphics[width=.49\textwidth]{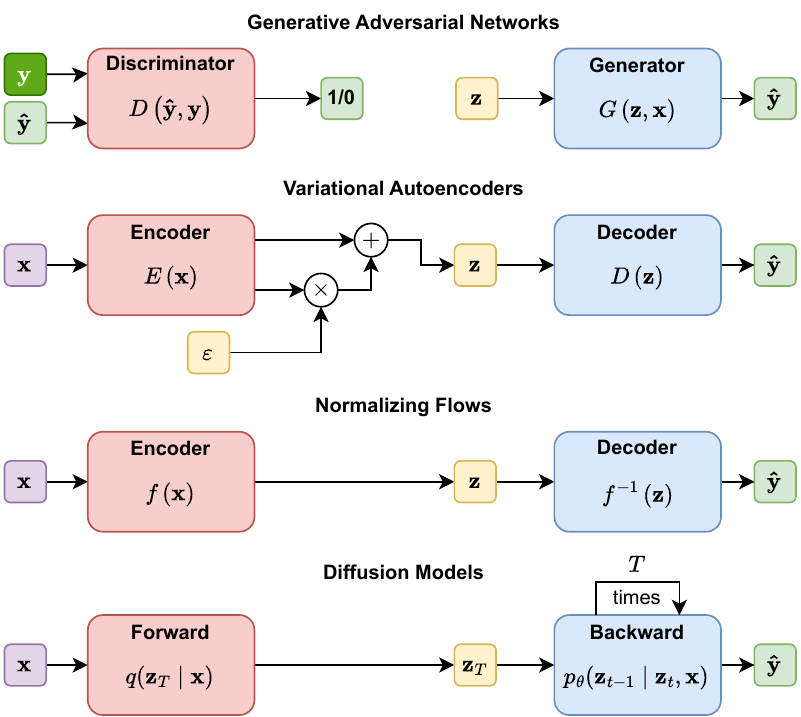}
        
        \caption{\label{fig:generativeModels}
        Conceptual overview of generative models (GANs, VAEs, NFs, and DMs). 
        }
    \end{center}
\end{figure}

\subsection{Relation between Diffusion Models}
As highlighted in the SDE section, we can describe both variations, namely SGMs, and DDPMs, with SDEs.
We can also showcase this close relationship by reformulating the optimization targets.
For DDPMs, we saw in \autoref{eq:vlb_ddpm} that
\begin{align}
    &\text{KL} ( q \left( \mathbf{z}_{0}, ..., \mathbf{z}_{T}\right) \| p_\theta \left( \mathbf{z}_{0}, ..., \mathbf{z}_{T}\right)) \nonumber\\
    &\stackrel{(ii)}{\geq} \mathbb{E} \left[ - \log p_\theta \left( \mathbf{z}_{0}\right)\right] + c \nonumber
\end{align}
is minimized. By reweighting the VLB, as Ho \textit{et al.} \cite{ho2020denoising} recommends for improved sample quality, we can further derive:
\begin{equation}
    \label{eq:ddpm_pos}
    \mathop{\mathbb{E}}_{\substack{t\sim\mathcal{U}(1,T) \\ \mathbf{z}_0 \sim q( \mathbf{z}_0) \\ \epsilon \sim \mathcal{N} \left( \mathbf{0}, \mathbf{I} \right)}} \left[ \lambda(t) \| \epsilon - \epsilon_\theta(\mathbf{z}_t, t)\|^2\right], \nonumber
\end{equation}
where $\lambda(t) > 0$ is a weighting function. If we now take the optimization target in \autoref{eq:sgm_lastOpt} of SGMs, which was
\begin{equation}
    \mathop{\mathbb{E}}_{\substack{t\sim\mathcal{U}(1,T) \nonumber\\ \mathbf{z}_0 \sim q( \mathbf{z}_0) \\ \epsilon \sim \mathcal{N} \left( \mathbf{0}, \mathbf{I} \right)}} \left[ \lambda(t) \| \epsilon + \sigma_t s_\theta(\mathbf{z}_t, t)\|^2\right] + c, \nonumber
\end{equation}
the connection between DDPMs and SGMs becomes clear once we set $ \epsilon_\theta(\mathbf{z}_t, t) = -\sigma_t s_\theta(\mathbf{z}_t, t)$.
As the constant $c$ is irrelevant for the optimization, we can see once again that there is a mathematical connection between DDPMs and SGMs.


\subsection{Relation to other Image SR Generative Models}
\label{sec:relation-gen-models}
Generative models in image SR differ primarily in how they approach the task of generating HR images from LR inputs and are illustrated in \autoref{fig:generativeModels}.
These differences stem from the underlying architecture and training objectives.
While they offer significant advantages, they come with a individual set of challenges, like training stability and computational costs.

\textbf{GAN:} One prominent category of generative models is Generative Adversarial Networks (GANs) \cite{gans}, which have demonstrated state-of-the-art performance in various vision-related tasks, including text-to-image synthesis \cite{frolov2021adversarial} and image super-resolution (SR) \cite{srgan}. 
GANs are known for their adversarial training, where a generator competes against a discriminator. 
Although DMs do not employ a discriminator, they utilize a similar adversarial training strategy by iteratively adding and removing noise to enable realistic data generation.
However, approaches with GANs often suffer from non-convergence, training instability, and high computational costs.
They require careful hyperparameter tuning due to the interplay between the generator and the discriminator.

\textbf{VAE:} 
Variational Autoencoders (VAEs) \cite{vae} are designed as autoencoders with a variational latent space, which is especially interesting in addressing the ill-posedness of image SR. 
The core objective of a VAE centers around establishing the variational lower bound of the log data likelihood, akin to the fundamental principle underlying DMs.
In a comparative context, one can consider DMs as a variation of VAEs but with a fixed VAE encoder responsible for perturbing the input data, while the VAE decoder resembles the backward diffusion process in DMs.
Still, unlike VAEs, which compress the input into smaller dimensions in the latent space, DMs often maintain the same spatial size. 

\textbf{ARM:} Autoregressive Models (ARMs) treat images as sequences of pixels and generate each pixel based on the values of previously generated pixels in a sequential manner \cite{guo2022lar}. 
The probability of the entire image is given as the product of conditional probability distributions for each individual pixel.
This makes ARMs computationally expensive for HR image generation.
Conversely, DMs generate data by gradually diffusing noise into an initial data sample and then reverse this process. 
Noise is diffused across the entire image simultaneously rather than sequentially.

\textbf{NF:} Normalizing Flows (NF) \cite{normalizing-flows} are a distinct category of generative models renowned for their capacity to represent data as intricate and complex distributions. 
Like DMs and VAEs, these models are optimized based on the log-likelihood of the data they generate. 
However, what sets NFs apart is their unique ability to learn an invertible parameterized transformation. 
Importantly, this transformation possesses a tractable Jacobian determinant, making it feasible to compute.
The concept of DiffFlow \cite{diff-flow} enters the picture as an innovative algorithm that marries the principles of DMs with those of NFs. 
This combination offers the promise of enhanced generative modeling capabilities. 
Yet, while promising, NFs are often considered challenging to train and can be computationally demanding \cite{papamakarios2021normalizing}.

\section{Improvements for Diffusion Models}
\label{sec:improvements}
In the broader research community, there are several ways to improve DMs for image generation, as presented, for example, by Karras \textit{et al.} \cite{karras2022elucidating}.
This section, however, focuses on enhancements particularly interesting for image SR: Efficient sampling and enhanced likelihood estimation.

\subsection{Efficient Sampling}
\label{sec:effSamp}
Efficient sampling refers to strategies that generate samples from noise more quickly, i.e., in fewer time steps, without compromising the quality of the produced image significantly.
For instance, a DDPM takes about 20 hours to sample 50,000 32x32 images, in contrast to a GAN's less than one minute on a Nvidia 2080 Ti GPU; for larger 256x256 images, this extends to nearly 1,000 hours \cite{vgg-face}.
Fortunately, the independence between training and inference schedules is often leveraged in image SR. 
For example, a model may undergo training with 1,000 time steps, but the subsequent inference phase may require only a fraction, i.e., 200 \cite{sr3-google, srdiff}. 
However, the broader community of DM research has made further attempts focusing on either training-based or training-free sampling.

\textbf{Training-based sampling} methods speed up data generation using a trained sampler that approximates the backward diffusion process instead of a traditional numerical solver. 
This process may be complete or partial. 
For example, Watson et al. \cite{fast-sampling-ddpm-1} developed a dynamic programming algorithm that identifies optimal inference paths using a fixed number of refinement steps, significantly reducing the computation required.
Diffusion Sampler Search \cite{fast-sampling-ddpm-2} offers another approach, optimizing fast samplers for pre-trained DMs by adjusting the Kernel Inception Distance. 
Another technique is truncated diffusion, which improves speed by prematurely ending the forward diffusion process \cite{truncated-diffusion-1, truncated-diffusion-2}. 
This early termination results in outputs that are not purely Gaussian noise, presenting computational challenges. 
These challenges are addressed using proxy distributions from pre-trained VAEs or GANs, which match the diffused data distribution and facilitate efficient backward diffusion.
Lastly, Knowledge distillation is also used to accelerate sampling. 
It involves transferring knowledge from a complex, slower sampler (the teacher model) to simpler, faster models (student models) \cite{distillation-guided-diffusion, knowledge-distillation}. 
As demonstrated by Salimans et al. \cite{progressive-distillation-diffusion}, this method progressively reduces the number of sampling steps, trading off a slight decrease in sample quality for increased speed.
Similarly, Xiao et al. \cite{Denoising-Diffusion-GANs} addressed the slow sampling issue associated with the Gaussian assumption in denoising steps, which is usually only effective for small step sizes. 
They proposed Denoising Diffusion GANs that use conditional GANs for the denoising steps, allowing for larger step sizes and faster sampling.
For image SR, an application for exploiting knowledge distillation can be found in AddSR \cite{xie2024addsr}. 
Similarly, YONOS-SR \cite{noroozi2024you} uses knowledge distillation, but instead of training faster samplers, they transfer different scaling task knowledge and use the training-free DDIMs for efficient sampling, which is presented in the next section.

\textbf{Training-free sampling} methods aim to speed up sampling by minimizing the number of discretization steps while solving the Stochastic Differential Equation (SDE) or Probability Flow Ordinary Differential Equation (ODE) \cite{ddim, score-based-fast-generation}.
Denoising Diffusion Implicit Models (DDIMs) \cite{ddim} introduced by Song \textit{et al.} generalizes the Markovian forward diffusion of DDPMs into non-Markovian ones. 
This generalization allows the DDIMs to learn a Markov chain to reverse the non-Markovian forward diffusion, resulting in higher sampling speeds with minimal loss in sample quality. 
Jolicoeur-Martineau \textit{et al.} \cite{score-based-fast-generation} have devised an efficient SDE solver with adaptive step sizes for the accelerated generation of score-based models. 
This method has been found to generate samples more rapidly than the Euler-Maruyama method without compromising sample quality.
Building upon DDIM and Jolicoeur-Martineau \textit{et al.}, the DPM-solver \cite{lu2022dpm}, inspired by the AnalyticalDPM \cite{bao2022analytic}, approximates the error prediction via Taylor expansion and thereby achieves efficient sampling by analytically resolving the linear component of the ODE solution instead of relying on generic black-box ODE solvers. 
This method significantly reduces the sampling steps to 10 to 20.
In a later work, the authors introduced an improved version with DPM-solver++ that essentially approximates the predicted image instead of the error \cite{lu2022dpm2}.
Lately, a more general formulation and extension of the DPM-solver++ was introduced by UniPC \cite{zhao2024unipc}.

\subsection{Improved Likelihood}
Log-likelihood improvement is directly coupled with enhancing the performance of various applications and methods, including but not limited to compression \cite{ho2019compression}, semi-supervised learning \cite{dai2017good}, and image SR. 
Given that DMs do not directly optimize the log-likelihood, e.g., SGMs utilize a weighted combination of score-matching losses, an objective that forms an upper bound on the negative log-likelihood needs to be optimized.
Song \textit{et al.} \cite{max-likelihood-diffusion} proposed a method called \textit{likelihood weighting} to address this need. This method minimizes the weighted combination of score matching losses for score-based DMs. 
A carefully chosen weighting function sets an upper bound on the negative log-likelihood in the weighted score-matching objective.
Upon minimization, this results in an elevation of the log-likelihood.
Kingma \textit{et al.} \cite{variational-dm} explored methods that simultaneously train the noise schedule and diffusion parameters to maximize the variational lower bound within Variational Diffusion Models.
Additionally, the Improved Denoising Diffusion Probabilistic Models (iDDPMs) proposed by Nichol and Dhariwal \textit{et al.} \cite{improved-ddpm} implement a cosine noise schedule. 
This gradually introduces noise into the input, contrasting with the linear schedules that tend to degrade the information quicker. 
Using the cosine noise schedule leads to better log-likelihoods and facilitates faster sampling. 

\section{Diffusion Models for Image SR}
\label{sec:diffusionSR}
\begin{figure}
    \begin{center}
        \includegraphics[width=.49\textwidth]{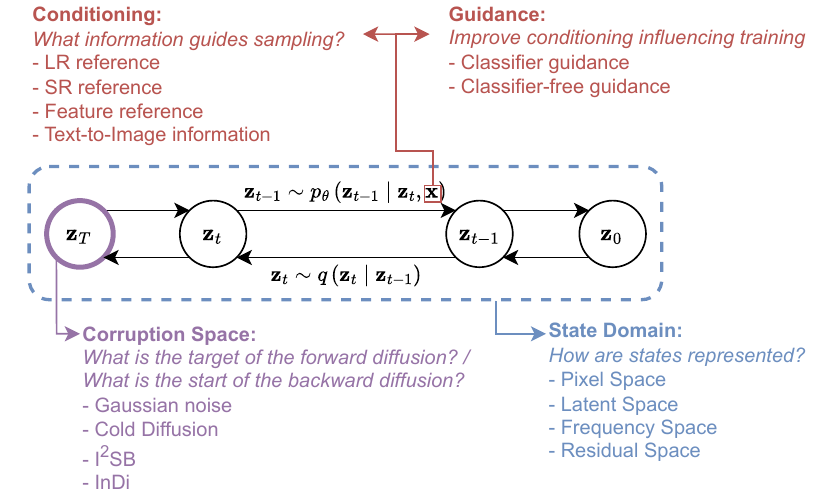}
        \caption{\label{fig:topology}
        Topology of this work. 
        Conditioning (\autoref{sec:conditioning}) leads the backward diffusion, whereas guidance (\autoref{sec:guidance}) is a training strategy to improve the incorporation of conditioning into DMs. The state domain (\autoref{sec:altDom}) describes the representation of states  $\mathbf{z}_t$. The corruption space (\autoref{sec:corrupt}) describes the target of the forward diffusion process or the start of the backward diffusion.
        }
    \end{center}
\end{figure}

So far, we introduced the theoretical framework of DMs. 
This section reviews practical applications and recent advances in image SR.
We will discuss concrete realizations of DMs, which are predominantly DDPMs. 
We then discuss guidance strategies to enhance conditioning usage, represent conditioning information in alternative state domains for DDPMs, and incorporate various conditioning methods.
Additionally, we explore SR-specific research areas, including corruption spaces, color-shifting, and architectural designs.
\autoref{fig:topology} provides a topological overview of this section.

\subsection{Concrete Realization of Diffusion Models}
While SGMs provide considerable design flexibility, the image SR trend leans towards DDPMs.
DDPMs benefit from a straightforward implementation, which reduces the entry barrier. 
It is a significant advantage, as it allows quicker development cycles and replication of results.
In addition, while the flexibility of SGMs is advantageous in creating customized solutions, it introduces design complexity due to the multitude of design variables that need to be considered. 
This poses a challenge in research settings, where rigorously evaluating the impact of each variable (e.g., different sampling algorithms) becomes cumbersome.
Moreover, the growing DDPM literature contributes to their popularity. 
As more studies adopt DDPMs, a virtuous cycle is created, where familiarity and proven effectiveness encourage further adoption. 

Among the pioneering DM efforts is SR3 \cite{sr3-google}, which concretely realizes DDPMs for image SR. 
Like typical for DDPMs, it adds Gaussian noise to the LR image until $\mathbf{z}_T \sim \mathcal{N} \left( \mathbf{0}, \mathbf{I} \right)$ and generates a target HR image $\mathbf{z}_0$ iteratively in $T$ refinement steps. 
SR3 employs the denoising model to predict the noise $\epsilon_t$.
The denoising model, $\varphi_\theta \left( \mathbf{x}, \mathbf{z}_t, \gamma_t \right)$, takes the LR image $\mathbf{x}$, the noise variance $\gamma_t$, and the noisy target image $\mathbf{z}_t$ as inputs.
With the prediction of $\epsilon_t$ provided by $\varphi_\theta$, we can reformulate \autoref{eq_sr3yt} to approximate $\mathbf{z}_0$ as follows:
\begin{equation}
\begin{split}
    & \mathbf{z}_t = \sqrt{\gamma_t} \cdot \mathbf{\hat{z}}_0  + \sqrt{1-\gamma_t} \cdot \varphi_\theta \left( \mathbf{x}, \mathbf{z}_t, \gamma_t \right) \\
    \iff & \mathbf{\hat{z}}_0 = \frac{1}{\sqrt{\gamma_t}} \cdot \left( \mathbf{z}_t - \sqrt{1-\gamma_t} \cdot \varphi_\theta \left( \mathbf{x}, \mathbf{z}_t, \gamma_t \right) \right)
\end{split}
\end{equation}
The substitution of $\mathbf{\hat{z}}_0$ into the posterior distribution to parameterize the mean of $p_\theta \left( \mathbf{z}_{t-1} | \mathbf{z}_t, \mathbf{x}\right)$ leads to:
\begin{equation}
\label{eq:sr3_parameterizedMean}
      \mu_\theta \left( \mathbf{x}, \mathbf{z}_t, \gamma_t \right)
      =  \frac{1}{\sqrt{\alpha_t}} \left[ \mathbf{z}_t - \frac{1 - \alpha_t}{\sqrt{1 - \gamma_t}} \cdot \varphi_\theta \left( \mathbf{x}, \mathbf{z}_t, \gamma_t \right)   \right]
\end{equation}
In SR3, the authors simplified the variance $\Sigma_\theta$ to $\left(1-\alpha_t\right)$ for ease of computation. Consequently, each refinement step with $\epsilon_t \sim \mathcal{N} \left( \mathbf{0}, \mathbf{I} \right)$ can be represented as:
\begin{equation}
    \mathbf{z}_{t-1} = \frac{1}{\sqrt{\alpha_t}} \left[ \mathbf{z}_t - \frac{1 - \alpha_t}{\sqrt{1 - \gamma_t}} \cdot \varphi_\theta \left( \mathbf{x}, \mathbf{z}_t, \gamma_t \right)   \right] + \sqrt{1-\alpha_t} \cdot \epsilon_t
\end{equation}
Concurrent work focused on a similar implementation of SR3 but shows different variations implementing the denoising model $\varphi_\theta \left( \mathbf{x}, \mathbf{z}_t, \gamma_t \right)$, which we will discuss later.
A notable mention is SRDiff \cite{srdiff}, published around the same time and follows a close realization of SR3.
The main distinction between SRDiff and SR3 is that SR3 predicts the HR image directly, whereas SRDiff predicts the residual information between the LR and HR image, i.e., the difference.
Thus, it has an alternative state domain, which will be discussed next.

\subsection{Guidance in Training}
\label{sec:guidance}
The backbone of diffusion-based image SR is the learning of conditional distributions \cite{ho2022cascaded, sr3-google}.
As such, the condition $\mathbf{x}$, e.g., the LR image, is integrated into the backward diffusion, i.e., $p_\theta \left( \mathbf{z}_{t-1} \mid \mathbf{z}_t, \mathbf{x} \right)$ for DDPMs or in $s_\theta(\mathbf{z}_{t}, \mathbf{x}, t)$ for SGMs/SDEs.
However, this simple formulation can result in a model that overlooks the conditioning. 
A principle known as guidance can mitigate this issue by controlling the weighting of the conditioning information at the expense of sample diversity. 
It can be categorized into classifier and classifier-free guidance.
To our knowledge, while effectively used for improving DMs, they have not been applied to image SR. 

\textbf{Classifier Guidance:}
Classifier guidance employs a classifier to guide the diffusion process by merging the score estimate of the DM with the gradients of the classifier during sampling \cite{dhariwal-diffusion-models-beat-gans}. 
This process is similar to low temperature or truncated sampling in BigGANs \cite{biggan} and facilitates a trade-off between mode coverage and sample fidelity.
The classifier is trained concurrently with the DM to predict the conditional information $\mathbf{x}$ from $\mathbf{z}_t$.
For weighting of the conditioning information, the score function becomes:
\begin{equation}
    \label{eq:class_guid}
    \nabla_{\mathbf{z}_t} \log{q(\mathbf{z}_t \mid \mathbf{x}}) = \nabla_{\mathbf{z}_t} \log{q(\mathbf{z}_t)} + \lambda \nabla_{\mathbf{z}_t} \log{q( \mathbf{x} \mid \mathbf{z}_t) },
\end{equation}
where $\lambda \in \mathbb{R}^+$ is a hyper-parameter for controlling the weighting.
The downside of this approach is its dependence on a learned classifier that can handle arbitrarily noisy inputs, a capability most existing pre-trained image classification models lack.

\textbf{Classifier-Free Guidance:}
Classifier-Free guidance aims to achieve similar results without a classifier \cite{ho2022classifier}. 
It modifies \autoref{eq:class_guid} into
\begin{equation}
    \nabla_{\mathbf{z}_t} \log{q(\mathbf{z}_t | \mathbf{x}}) = (1-\lambda) \nabla_{\mathbf{z}_t} \log{q(\mathbf{z}_t)} + \lambda \nabla_{\mathbf{z}_t} \log{q( \mathbf{z}_t \mid \mathbf{x})}.
\end{equation}
As a result, we have a standard unconditional DM and a conditional DM that has the score estimate $\nabla_{\mathbf{z}_t} \log{q( \mathbf{z}_t \mid \mathbf{x})}$. 
The unconditional DM remains when $\lambda = 0$, and for $\lambda = 1$, it aligns with the vanilla formulation of the conditional DM.
The interesting scenario arises when $\lambda > 1$, where the DM prioritizes conditional information and moves away from the unconditional score function, thus reducing the likelihood of generating samples disregarding conditioning information. 
However, the major downside of this approach is its computational cost for training two separate DMs. 
This can be mitigated by training a single conditional model and substituting the conditioning information with a null value in the unconditional score function \cite{luo2022understanding}.

\subsection{State Domains}
\label{sec:altDom}
\begin{figure}
    \begin{center}
        \includegraphics[width=.49\textwidth]{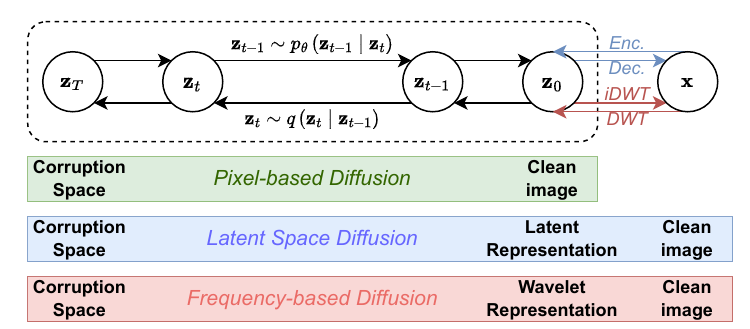}
        
        \caption{\label{fig:latentSpace}
        Overview of state domains. The green bar shows the vanilla DM operating in pixel space. The blue bar shows the exploit of the latent space domain via Autoencoders. The red bar shows the application of DMs in the wavelet domain.  
        }
    \end{center}
\end{figure}

So far, we have discussed methods that operate directly on the pixel space.
This section introduces different methods that map the input into alternative state domains: latent, frequency, and residual space. Apart from particular challenges arising from the alternative state domain, these methods incur an additional step that maps the pixel domain into their own, as illustrated in \autoref{fig:latentSpace}.

\textbf{Latent Space:}
Models like SR3 \cite{sr3-google}, and SRDiff \cite{srdiff} have achieved high-quality SR results by operating in the pixel domain. 
However, these models are computationally intensive due to their iterative nature and the high-dimensional calculations in RGB space. 
To reduce computational demands, one can move the diffusion process into the latent space of an autoencoder \cite{kim2024arbitrary}.
The first of this kind was the Latent Score-based Generative Models (LSGMs) by Vadhat \textit{et al.} \cite{score-based-ldm}.
It is a regular SGM that operates in the latent space of a VAE and, by pre-training the VAE, achieves even faster sampling speeds.
It yields comparable and better results than DMs operating in the pixel domain while being faster.
Building upon LSGMs, Rombach \textit{et al.} introduced the Latent Diffusion Model (LDM) \cite{wang2023exploiting, rombach-latent-diffusion-models}, which also performs diffusion in a low-dimensional latent space of an autoencoder. 
In contrast to LSGM, LDM utilizes a DDPM and an autoencoder that is pre-trained, like the VQ-GAN \cite{esser2021taming}, and is not jointly trained with the denoising network.
This approach significantly lowers resource requirements without compromising performance. 
Due to the decoupled training, it requires very little regularization of the latent space and allows the reuse of latent representations across multiple models.
Improving upon LDMs is REFUSION (image REstoration with difFUSION models) \cite{luo2023image} by Luo \textit{et al.}, which differs in two aspects:
First, it uses a U-Net that contains skip connections from the encoder to the decoder, which provides the decoder with additional details.
Moreover, it introduces Nonlinear Activation-Free blocks (NAFBlocks) \cite{chen2022simple}, replacing all non-linear activations with an element-wise operation that splits feature channels into two parts and multiplies them to produce one output.
Secondly, they train their U-Net with a latent-replacing training strategy, which partially replaces the latent representation with either the encoded LR or HR image for reconstruction training.
Similarly, Chen \textit{et al.} \cite{chen2023hierarchical} improve the architectural aspects of LDMs and propose a two-stage strategy called the Hierarchical Integration Diffusion Model (HI-Diff). 
In the first stage, an encoder compresses the ground truth image to a highly compact latent space representation, which has a much higher compression ratio than LDM. 
As a result, the computational burden of the DM, which refines multi-scale latent representations, is much more reduced.
The second stage is a vision transformer-based autoencoder, which incorporates the latent representations of the first stage during the downsampling process via Hierarchical Integration Modules (HIM), a cross-attention fusion module.

\textbf{Frequency Space:}
Wavelets provide a novel outlook on SR \cite{10041995, 10.1007/978-3-031-44210-0_19}.
The conversion from the spatial to the wavelet domain is lossless and offers significant advantages as the spatial size of an image can be downsized by a factor of four, thereby allowing faster diffusion during the training and inference stages.
Moreover, the conversion segregates high-frequency details into distinct channels, facilitating a more concentrated and intentional focus on high-frequency information, offering a higher degree of control \cite{guo2017deep}.
Besides, it can be conveniently incorporated into existing DMs as a plug-in feature.
The diffusion process can interact directly with all wavelet bands as proposed in DiWa \cite{moser2023waving} or specifically target certain bands while the remaining bands are predicted via standard CNNs. 
For instance, WaveDM \cite{huang2023wavedm} modifies the low-frequency band, whereas WSGM \cite{guth2022wavelet} or ResDiff \cite{shang2023resdiff} conditions the high-frequency bands relative to the low-resolution image.
Altogether, the wavelet domain presents a promising avenue for future research. 
It provides potential for significant performance acceleration while maintaining, if not enhancing, the quality of SR results.
%
%

\textbf{Residual Space:}
SRDiff \cite{srdiff} was the first work that advocated for shifting the generation process into the residual space, i.e., the difference between the upsampled LR and the HR image. 
This enables the DM to focus on residual details, speeds up convergence, and stabilizes the training \cite{10041995, 10.1007/978-3-031-44210-0_19}. 
Whang \textit{et al.} \cite{deblur-diff} also employs residual predictions as a fundamental component of their predict-and-refine approach for image deblurring.
However, unlike SRDiff, they provide a SR prediction with a CNN instead of the bilinear upsampled LR and predict the residuals between the SR prediction and the HR ground truth with their DM.
An improvement is presented by ResDiff \cite{shang2023resdiff}, which additionally incorporates the SR prediction and its high-frequency information during the backward diffusion for better guidance. 
In a different vein, Yue \textit{et al.} \cite{yue2023resshift} presents ResShift. 
This technique constructs a Markov chain of transformations between HR and LR images by manipulating the residual between them. 
Thus, instead of just adding Gaussian noise with zero mean in the forward process, the residual is also added as the mean of the noise sampling during training.
This novel approach substantially enhances sampling efficiency, i.e., only 15 sampling steps.

\subsection{Conditioning Diffusion Models}
\label{sec:conditioning}
DMs depend on conditioning information to guide the sampling process toward a reasonable HR prediction. 
One common strategy is to use the LR image during the backward diffusion.
This section reviews various alternative methods for integrating conditioning information into backward diffusion.

\textbf{Low Resolution Reference:}
High-quality SR predictions can be achieved through a straightforward channel concatenation \cite{saharia2022photorealistic}. 
The LR image is concatenated with the denoised result from time step $t - 1$ and serves as the conditioning input for noise prediction at time step $t$.
In contrast, Iterative Latent Variable Refinement (ILVR) by Choi \textit{et al.} \cite{choi2021ilvr} conditions the generative process of an unconditional LDM \cite{rombach-latent-diffusion-models}. 
This approach offers the advantage of shorter training times, as it leverages a pre-trained DM.
To integrate conditioning information, the low-frequency components of the denoised output are replaced with their corresponding counterparts from the LR image. 
Thus, the latent variable is aligned with a provided reference image at each generation process stage, ensuring precise control and adaptation during generation.

\textbf{Super-Resolved Reference:}
An alternative to conditioning the denoising on the LR image involves learned priors from pre-trained SR models to predict a reference image. 
E.g., CDPMSR \cite{niu2023cdpmsr} conditions the denoising process with a predicted SR reference image obtained using existing and standard SR models.
ResDiff \cite{shang2023resdiff}, on the other hand, leverages a pre-trained CNN to predict a low-frequency, content-rich image that includes partial high-frequency components.
This image guides the noise towards the residual space, offering an alternative means of conditioning the generative process.

\begin{figure}
    \begin{center}
        \includegraphics[width=.49\textwidth]{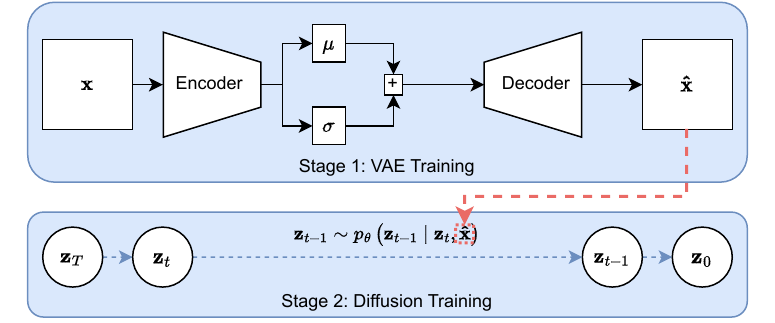}
        
        \caption{\label{fig:diffuseVAE}
        Overview of DiffuseVAE. The two-stage approach employs a VAE (first stage), which generates variational prediction as a condition for the DM (second stage).
        }
    \end{center}
\end{figure}

\begin{table}
\caption{Results for 4× SR of general images on DIV2K val. 
Note that EDSR, FxSR-PD, CAR, and RRDB are regression-based methods that generally produce better PSNR and SSIM scores than generative approaches \cite{sr3-google}. 
}
\center
\small
\begin{tabular}{l c c c }
\toprule
\textbf{Methods} & \textbf{PSNR} $\uparrow$ & \textbf{SSIM} $\uparrow$& \textbf{LPIPS} $\downarrow$ \\ 
\midrule
Bicubic & 26.70 & 0.77 & 0.409  \\
EDSR \cite{lim2017enhanced} & 28.98 & 0.83 & 0.270  \\
FxSR-PD \cite{park2022flexible} & 29.24 & 0.84 & 0.239 \\
RRDB \cite{esrgan} & 29.44 & 0.84 & 0.253  \\ 
CAR \cite{sun2020learned} & 32.82 & 0.88 & -  \\ 
\midrule
RankSRGAN \cite{zhang2019ranksrgan} & 26.55 & 0.75 & 0.128  \\ 
ESRGAN \cite{esrgan} & 26.22 & 0.75 & 0.124  \\
SRFlow \cite{lugmayr2020srflow} & 27.09 & 0.76 & \underline{0.120}  \\
\midrule
SRDiff \cite{srdiff} & 27.41 & 0.79 & 0.136  \\
IDM \cite{improved-ddpm} &  27.59 & 0.78 & - \\
DiWa \cite{moser2023waving}  & 28.09 &  0.78 & 0.104\\
\bottomrule
\end{tabular}
\label{tab:div2k_results}
\end{table}

\begin{table}
\caption{PSNR and SSIM comparison on CelebA-HQ face SR 16$\times$16 $\rightarrow$ 128$\times$128. 
Consistency measures MSE ($\times 10^{-5}$) between LR inputs and the downsampled SR outputs. }
\center
\small
\begin{tabular}{l c c c }
\toprule
\textbf{Methods} & \textbf{PSNR} $\uparrow$ & \textbf{SSIM} $\uparrow$& \textbf{Consistency} $\downarrow$ \\ 
\midrule
PULSE \cite{pulse} & 16.88 & 0.44 & 161.1 \\
FSRGAN \cite{chen2018fsrnet} & 23.01 & 0.62 & 33.8 \\
SR3 (regression) \cite{sr3-google} & 23.96 & 0.69 & 2.71 \\
\midrule
SR3 (diffusion)  \cite{sr3-google} & 23.04 & 0.65 & 2.68 \\
DiWa \cite{moser2023waving} & 23.34 & 0.67 & - \\
IDM \cite{improved-ddpm} & 24.01 & 0.71 & 2.14 \\
\bottomrule
\end{tabular}
\label{tab:celeba}
\end{table}

Pandey \textit{et al.} \cite{pandey2022diffusevae} introduced an exciting idea of varying predicted conditions with DiffuseVAE as illustrated in \autoref{fig:diffuseVAE}.
This approach integrates the stochastic predictions generated by a VAE as conditioning information for the DM, capitalizing on the advantages offered by both models. 
They use a two-stage approach called the \textit{generator-refiner} framework.
In the first stage, a VAE is trained on the training data. 
In the subsequent stage, the DM is conditioned using varying, often blurred, reconstructions generated by the VAE.
The essential advantage of this method lies in the diversity in the generated samples, which is defined within the lower-dimensional latent space of the VAE. 
This characteristic creates a more favorable balance between sampling speed and sample quality. 
It is advantageous in scenarios where multiple predictions are required, similar to the use cases for Normalizing Flows.

\textbf{Feature Reference:}
%
Another avenue for conditioning involves relevant features extracted from pre-trained networks.
SRDiff \cite{srdiff} leverages a pre-trained encoder to encode LR image features at each step of the backward diffusion. 
These features serve as guidance, aiding in the generation of higher-resolution outputs.  
Implicit DMs (IDMs) \cite{improved-ddpm} take a different approach by conditioning their denoising network with a neural representation, which enables the learning of a continuous representation at various scales. 
They encode the image as a function within continuous space and seamlessly integrate it into the DM.
These extracted features are adapted to multiple scales and are used across multiple layers within the DMs.
To comprehensively understand the performance differences between these approaches, comparisons can be found in \autoref{tab:celeba} and \autoref{tab:div2k_results}.
Recently, DeeDSR was introduced \cite{bi2024deedsr}, which incorporates degradation-aware features extracted from the LR image to guide the diffusion process of a LDM \cite{wang2023exploiting}.

\textbf{Text-to-Image Information:}
By incorporating conditioning information that goes beyond the LR image (e.g., its SR prediction, direct concatenation of the LR image, or its feature representation), one can add Text-To-Image (T2I) information.
The incorporation of T2I information proves advantageous as it allows the usage of pre-trained T2I models. 
These models can be fine-tuned by adding specific layers or encoders tailored to the SR task, facilitating the integration of textual descriptions into the image generation process. 
This approach enables a richer source of guidance, potentially improving image synthesis and interpretation in SR tasks.
Wang \textit{et al.} have put this concept into practice with StableSR \cite{wang2023exploiting}.
Central to StableSR is a time-aware encoder trained in tandem with a frozen Stable DM, essentially a LDM. 
This setup seamlessly integrates trainable spatial feature transform layers, enabling conditioning based on the input image.
To further augment the flexibility of StableSR and achieve a delicate balance between realism and fidelity, they introduce an optional controllable feature wrapping module. 
This module accommodates user preferences, allowing for fine-tuned adjustments based on individual requirements. 
The inspiration for this feature comes from the methodology introduced in CodeFormer \cite{zhou2022towards}, which enhances the versatility of StableSR in catering to diverse user needs and preferences.
Likewise, Yang \textit{et al.} introduce a method known as Pixel-Aware Stable Diffusion (PASD) \cite{yang2023pixel}. 
PASD takes conditioning a step further by incorporating text embeddings of the LR input using a CLIP text encoder \cite{radford2021learning} and its feature representation.
This approach augments the model's ability to generate images by incorporating textual information, thus allowing for more precise and context-aware image synthesis. 
Comparisons between PASD and other approaches can be found in \autoref{tab:t2i}, demonstrating the impact of this text-based conditioning on image SR results.
A similar concurrent work can be found with SeeSR \cite{wu2023seesr}.
XPSR \cite{qu2024xpsr} extends this idea by fusing different levels of semantic text encodings (high-level: the content of the image; low-level: the perception of overall quality, sharpness, noise level, and other distortions about the LR image).

\begin{table}
\caption{Results for 4× SR of general images on resized DIV2K val ($128 \times 128 \rightarrow 512 \times 512$). 
}
\center
\small
\begin{tabular}{l c c c }
\toprule
\textbf{Methods} & \textbf{PSNR} $\uparrow$ & \textbf{SSIM} $\uparrow$& \textbf{LPIPS} $\downarrow$ \\ 
\midrule
BSRGAN \cite{zhang2021designing} & 23.41 & 0.61 & 0.426  \\
Real-ESRGAN \cite{wang2021real} & 23.15 & 0.62 & 0.403  \\ 
LDL \cite{liang2022details} & 22.74 & 0.62 & 0.416 \\
FeMaSR \cite{chen2022femasr} & 21.86 & 0.54 & 0.410  \\ 
SwinIR-GAN \cite{liang2021swinir} & 22.65 & 0.61 & 0.406  \\ 
\midrule
LDM \cite{rombach-latent-diffusion-models} & 21.48 & 0.56 & 0.450  \\ 
SD Upscaler \cite{rombach-latent-diffusion-models} & 21.21 & 0.55 & 0.430  \\
StableSR \cite{wang2023exploiting} & 20.88 & 0.53 & 0.438  \\
PASD \cite{yang2023pixel} & 21.85 & 0.52 & 0.403  \\
\bottomrule
\end{tabular}
\label{tab:t2i}
\end{table}

\begin{figure}[!t]
    \begin{center}
        \includegraphics[width=.49\textwidth]{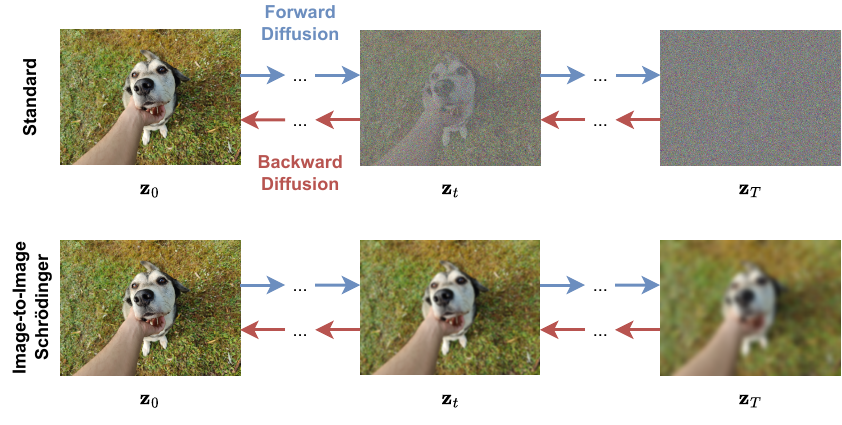}
        
        \caption{\label{fig:isb}
        Comparison of the standard corruption space and I$^2$SB. Instead of injecting noise to the clean image (initial state $\mathbf{z}_0$), the final state $\mathbf{z}_T$ is the degraded image.
        }
    \end{center}
\end{figure}

\subsection{Corruption Space}
\label{sec:corrupt}
Karras \textit{et al.} \cite{karras2022elucidating} identified three pillars of DMs: the noise schedule, the network parameterization, and the sampling algorithm.
Recently, many authors argued to consider also different types of corruption instead of pure Gaussian noise used during forward diffusion like Soft Score Matching \cite{daras2022soft}, i.e., the starting point for backward diffusion or the target for the forward diffusion $\mathbf{z}_T$.
Soft Score Matching directly incorporates the filtering process within the SGM, training the model to predict a clean image. 
Upon corruption, this predicted image aligns with the diffused observation.
Note that $\mathbf{z}_T$ may be represented differently due to alternative state domains (e.g., latent, frequency, or residual).
Cold Diffusion \cite{bansal2022cold} presents another ingenious way of modifying the corruption space for DDPMs.
It shows that the generative capability is not strongly dependent on the choice of image degradation.
It reveals new experimental types of diffusion besides Gaussian noise can be effectively used, like animorphosis (i.e., human faces iteratively degrading to animal faces).
The Image-to-Image Schr{\"o}dinger Bridge (I$^2$SB) goes in a similar direction but does not impose any assumptions on the underlying prior distributions \cite{liu20232}. 
In its diffusion process, the clean image represents the initial state, while the degraded image is the final state in both forward and backward diffusion processes.
This is notable for its ability to provide a transparent and traceable path from a degraded image to its clean version, as illustrated in \autoref{fig:isb}.
Consequently, it provides enhanced interpretability since the process between degraded and clean images is directly addressed, which is not commonly present in many DMs.
Another benefit is its higher efficiency in backward diffusion since it requires fewer steps (often between 2 and 10) to achieve comparable performance.
Its conditionality, however, limits its use specifically to paired data during training, which is unsuitable for unsupervised SR.
While Cold Diffusion and I$^2$SB show promising results for image restoration, an extensive and more detailed quantitative analysis of different corruption types for image SR remains an exciting and open research avenue.
Another avenue for alternative corruption space is presented by Inversion by Direct Iteration (InDI) \cite{delbracio2023inversion}.
InDI delineates a direct mapping strategy, efficiently bridging the gap between the two quality spaces without the iterative refinement typically required by conventional diffusion processes.
The intrinsic flexibility and the direct mapping capability of InDI propose intriguing possibilities for enhancing image quality, suggesting a potent avenue for research exploration.
The potential integration of InDI's principles with those of conditional DMs could offer substantial advancements in the field of image SR. 
A detailed examination and discussion of InDI within the broader scope of diffusion-based image enhancement could yield valuable insights and contribute significantly to the ongoing development of generative models in image processing.

\begin{figure}[!t]
    \begin{center}
        \includegraphics[width=.3\textwidth]{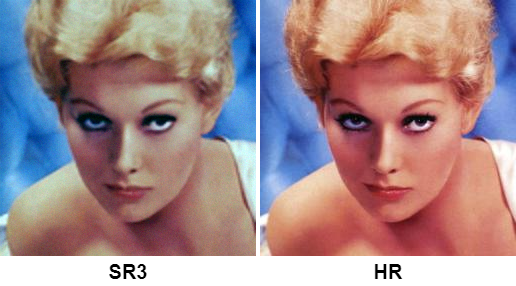}
        
        \caption{\label{fig:color}
        Example of color shifting produced by vanilla SR3 in a $64\times64 \rightarrow 256\times256$ setting when trained with reduced batch size (8 instead of 256). 
        }
    \end{center}
\end{figure}

\subsection{Color Shifting}
\label{sec:colorShift}
As a result of high computational costs, DMs can occasionally suffer from color shifting when limited hardware necessitates smaller batch sizes or shorter learning periods \cite{choi2022perception}.
An example with SR3 is shown in \autoref{fig:color}.
As presented by StableSR, a straightforward modification can address this issue by performing color normalization by adjusting the mean and variance with those of the LR image on the generated image \cite{wang2023exploiting}.
Mathematically, it gives the following equation:
\begin{equation}
    \mathbf{\hat{z}}_0 = \frac{\mathbf{z}_0^c - \mu_{\mathbf{z}_0}^c}{\sigma^c_\mathbf{\mathbf{z}_0}} \cdot \sigma_\mathbf{x}^c + \mu_\mathbf{x}^c,
\end{equation}
\noindent
where $c \in \{ r, g, b\}$ denotes the color channel, and $\mu_{\mathbf{z}_0}^c$ and $\sigma^c_\mathbf{\mathbf{z}_0}$ (or $ \mu_\mathbf{x}^c$ and $\sigma_\mathbf{x}^c$) are the mean and standard variance from the $c$-th channel of the predicted image $\mathbf{z}_0$ (or the input image $\mathbf{x}$), respectively.
You Only Diffuse Areas (YODA) \cite{moser2023yoda}, which targets diffusion on important image areas more frequently through time-dependent masks generated with DINO \cite{caron2021emerging}, also mitigates the color shift effect for image SR.
This suggests that properly defined architecture and diffusion design are crucial to omit this effect.
Further analysis of why this effect emerges must be obtained in future work.

\subsection{Architecture Designs for Denoising}

The design of the denoising model in DMs offers a range of options. 
The majority of DMs adopt the use of U-Net, as noted in most literature \cite{biggan}. 
SR3 \cite{sr3-google}, for instance, employs residual blocks from BigGAN \cite{biggan} and re-scales skip connections by a factor of $\frac{1}{\sqrt{2}}$. 
SRDiff takes a similar approach \cite{srdiff}, although it opts for vanilla residual blocks without the re-scaling of skip connections and uses a LR encoder to incorporate the information of the LR image during the backward diffusion.
Whang \textit{et al.} \cite{deblur-diff} exploit an initial predictor to combine the strengths of deterministic image SR models and DMs. 
It has the advantage that the DM only needs to learn the residuals that the deterministic image SR model (initial predictor) fails to predict, hence simplifying the learning target. Additionally, the removal of self-attention, positional encodings and group normalization from the SR3 U-Net enables their model to support arbitrary resolutions.
An initial predictor is also employed in the wavelet-based approach DiWa \cite{moser2023waving}. 
Moreover, wavelet SR models, such as DWSR \cite{guo2017deep} – a simple sequence of convolution layers of depth 10 – are utilized for denoising prediction in the wavelet domain. 
In WaveDM \cite{huang2023wavedm}, a deterministic U-Net predictor is used for the high-frequency band, while diffusion is applied in the low-frequency band.

Latent Diffusion Models proposed by Rombach \textit{et al.} \cite{rombach-latent-diffusion-models} use a VQ-GAN \cite{esser2021taming} autoencoder in the latent space. 
For DiffIR \cite{xia2023diffir}, multiple variations of state-of-the-art Vision Transformers are employed \cite{vaswani2017attention, dosovitskiy2020image, liang2021swinir}.
Another common practice is pre-training deterministic components, as seen in models like SRDiff \cite{srdiff} or DiffIR \cite{xia2023diffir}. 
Overall, the potential ways to design a denoising network are infinite, generally drawing inspiration from advancements made in general computer vision. 
The optimal denoising networks will vary based on the task, and the development of new models is anticipated.

\begin{table*}
\center
\caption{\label{tab:zs_comparison}Comparison of zero-shot methods. Data in bold represents the best performance. Second-best is underlined. Values derived from Li \textit{et al.} \cite{li2023diffusion}.}
\small
\begin{tabular}{l | c c c | c c c | c | c}
\toprule
\multirow{2}{*}{\textbf{Methods}} & \multicolumn{3}{c}{\textbf{ImageNet 1K}} & \multicolumn{3}{c}{\textbf{CelebA 1K}} & \textbf{Time} & \textbf{Flops}\\
& \textbf{PSNR} $\uparrow$ & \textbf{SSIM} $\uparrow$& \textbf{LPIPS} $\downarrow$ & \textbf{PSNR} $\uparrow$ & \textbf{SSIM} $\uparrow$& \textbf{LPIPS} $\downarrow$ & [s/image] & [G]\\ 
\midrule
Bicubic & 25.36 & 0.643 & 0.27 & 24.26 & 0.628 & 0.34 & - & - \\
ILVR \cite{choi2021ilvr} & \underline{27.40} & \textbf{0.871} & 0.21 & \underline{31.59} & 0.878 & 0.22 & 41.3 & 1113.75 \\
SNIPS \cite{kawar2021snips} & 24.31 & 0.684 & 0.21 & 27.34 & 0.675 & 0.27 & 31.4 & - \\
DDRM \cite{kawar2022denoising} & 27.38 & \underline{0.869} & 0.22 & \textbf{31.64} & \textbf{0.946} & 0.19 & \underline{10.1} & 1113.75 \\
DPS \cite{chung2022diffusion} & 25.88 & 0.814 & \underline{0.15} & 29.65 & 0.878 & 0.18 & 141.2 & 1113.75 \\
DDNM \cite{wang2022zero} & \textbf{27.46} & \textbf{0.871} & \underline{0.15} & \textbf{31.64} & \underline{0.945} & \textbf{0.16} & 15.5 & 1113.75 \\
GDP \cite{fei2023generative} & 26.51 & 0.832 & \textbf{0.14} & 28.65 & 0.876 & \underline{0.17} & \textbf{3.1} & 1113.76 \\
\bottomrule
\end{tabular}
\end{table*}

\section{Diffusion-based Zero-Shot SR}
Zero-shot image SR aims to develop methods that do not depend on prior image examples or training \cite{shocher2018zero, 10041995}. 
Typically, these methods harness the inherent redundancy within a single image for improvement. 
They often leverage pre-trained DMs for generation, incorporating LR images as conditions during the sampling process, in contrast to other conditioning methods discussed earlier \cite{li2024omnissr}. 
Additionally, they differ from guidance-based methods, where conditioning information is used to weight the training of a DM from scratch.
A recent study by Li \textit{et al.} \cite{li2023diffusion} categorizes diffusion-based methods into projection-based, decomposition-based, and posterior estimation, which are introduced in this section.
The discussed methods are compared in \autoref{tab:zs_comparison}.

\subsection{Projection-Based}
Projection-based methods aim to extract inherent structures or textures from LR images to complement the generated images at each step and to ensure data consistency.
An illustrative example of a projection-based method in the realm of inpainting tasks is RePaint \cite{lugmayr2022repaint}. 
In RePaint, the diffusion process is selectively applied to the specific area requiring inpainting, leaving the remaining image portions unaltered.
Taking inspiration from this concept, YODA \cite{moser2023yoda} applies a similar technique, but for image SR. 
YODA incorporates importance masks derived from DINO \cite{caron2021emerging} to define the areas for diffusion during each time step, but it is not a zero-shot approach.

One zero-shot method is ILVR \cite{choi2021ilvr}, which projects the low-frequency information from the LR image to the HR image, ensuring data consistency and establishing an improved DM condition.
A more sophisticated method is Come-Closer-Diffuse-Faster (CCDF) \cite{chung2022ccdf}, which modifies the unified projection method to SR as follows:
\begin{align}
    \mathbf{\hat{z}}_{t-1} &= \text{f}(\mathbf{z}_{t}, t) + \text{g}(\mathbf{z}_{t}, t) \cdot \varepsilon_t \\
    \mathbf{z}_{t-1} &= (\mathbf{I}-\mathbf{P}) \cdot \mathbf{\hat{z}}_{t-1} + \mathbf{\hat{x}}, \quad \mathbf{\hat{x}} \sim q(\mathbf{z}_t | \mathbf{z}_0 = \mathbf{x}),
\end{align}
\noindent
where $\text{f}, \text{g}$ depend on the type of DMs, $\mathbf{P}$ is the degradation process of the LR image, and $\mathbf{\hat{x}}$ is the LR image with the added and time-dependent noise.

\subsection{Decomposition-Based}
Decomposition-based methods view image SR tasks as a linear reverse problem similar to \autoref{eq:degradation}:
\begin{equation}
    \mathbf{x} = \mathbf{A} \mathbf{y} + b,
\end{equation}
where $\mathbf{A}$ is the degradation operator and $b$ contaminating noise.
Among the earliest decomposition-based methods, we find SNIPS \cite{kawar2021snips} and its subsequent work DDRM \cite{kawar2022denoising}. 
These methods employ diffusion in the spectral domain, enhancing SR outcomes.
To achieve this, they apply singular value decomposition to the degradation operator $\mathbf{A}$, thereby facilitating a spectral-domain transformation that contributes to their improved SR results.

The Denoising Diffusion Null-space Model (DDNM) represents another decomposition-based zero-shot approach applicable to a broad range of linear IR problems \cite{wang2022zero} beyond image SR to tasks like colorization, inpainting, and deblurring \cite{wang2022zero}. 
It leverages the range-null space decomposition methodology \cite{schwab2019deep, wang2023gan} to tackle diverse IR challenges effectively.
DDNM approaches the problem by reconfiguring \autoref{eq:degradation} as a linear reverse problem, although it is essential to note that this approach differs from SNIPS and DDRM in that it operates in a noiseless context:
\begin{equation}
    \mathbf{x} = \mathbf{A}\mathbf{y},
\end{equation}
with $\mathbf{y}\in \mathbb{R}^{D \times 1}$ as the linearized HR image and $\mathbf{x} \in \mathbb{R}^{d \times 1}$ the linearized degraded image.
Furthermore, it has to conform to the following two constraints:
\begin{equation}
    \textit{Consistency}: \quad \mathbf{A}\mathbf{\hat{y}} \equiv \mathbf{x},\quad\quad\textit{Realness}: \quad \mathbf{\hat{y}} \sim p(\mathbf{y}),
    \label{eq:consistency}
\end{equation}
with $p(\mathbf{y})$ as the distribution of ground-truth images and $\mathbf{\hat{y}}$ the predicted image.
The range-null space decomposition allows constructing a general solution for $\mathbf{\hat{y}}$ in the form of:
\begin{equation}
    \mathbf{\hat{y}}=\mathbf{A^{\dagger}}\mathbf{x} + (\mathbf{I} - \mathbf{A^{\dagger}}\mathbf{A})\bar{\mathbf{y}},
\end{equation}
with $\mathbf{A^{\dagger}}\in\mathbb{R}^{D\times d}$ the pseudo-inverse that satisfies $\mathbf{A}\mathbf{A^{\dagger}\mathbf{A}}\equiv\mathbf{A}$. 
Our goal is to find a proper $\bar{\mathbf{y}}$ that generates the null-space $(\mathbf{I} - \mathbf{A^{\dagger}}\mathbf{A})\bar{\mathbf{y}}$ and agrees with the range-space $\mathbf{A^{\dagger}}\mathbf{x}$ that also fulfills realness in \autoref{eq:consistency}.

\begin{figure}
    \begin{center}
        \includegraphics[width=.49\textwidth]{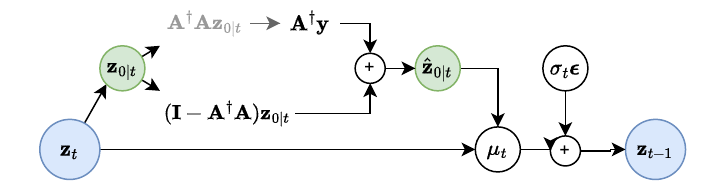}
        
        \caption{\label{fig:ddnm}
        Overview of DDNM \cite{wang2022zero}. It utilizes the range-null space decomposition to construct a general solution for multiple tasks, such as image SR, colorization, inpainting, and deblurring.
        }
    \end{center}
\end{figure}

DDNM derives clean intermediate states, denoted as $\mathbf{z}_{0|t}$, for the range-null space decomposition from $\mathbf{z}_{0}$ at time-step $t$. This is achieved through the equation:
\begin{equation}
    \mathbf{z}_{0|t} = \frac{1}{\sqrt{\Bar{\alpha}_{t}}}	\left( \mathbf{z}_{t} - \epsilon_{\theta}(\mathbf{z}_{t},t)\sqrt{1-\Bar{\alpha}_{t}} \right)
\end{equation}
with $\epsilon_t=\epsilon_\theta(\mathbf{z}_{t},t)$.
To produce a $\mathbf{z}_{0}$ that fulfills the equation $\mathbf{A}\mathbf{z}_{0}\equiv \mathbf{x}$, the model leaves the null-space unaltered while setting the range-space as $\mathbf{A}^{\dagger}\mathbf{y}$. This generates a rectified estimation, $\hat{\mathbf{z}}_{0|t}$, defined by:
\begin{equation}
    \hat{\mathbf{z}}_{0|t}=\mathbf{A^{\dagger}}\mathbf{x} + (\mathbf{I} - \mathbf{A^{\dagger}}\mathbf{A})\mathbf{z}_{0|t}.
    \label{eq:ndm core}
\end{equation}
Finally, $\mathbf{z}_{t-1}$ is derived by sampling from $p(\mathbf{z}_{t-1}|\mathbf{z}_{t},\hat{\mathbf{z}}_{0|t})$:
\begin{equation}
   \mathbf{z}_{t-1} = \frac{\sqrt{\Bar{\alpha}_{t-1}}\beta_{t}}{1-\Bar{\alpha}_{t}}\hat{\mathbf{z}}_{0|t}+ \frac{\sqrt{\alpha_{t}}(1-\Bar{\alpha}_{t-1})}{1-\Bar{\alpha}_{t}}\mathbf{z}_{t} + \sigma_{t}\boldsymbol{\epsilon}, \quad \boldsymbol{\epsilon}\sim \mathcal{N}(0,\mathbf{I}),
    \label{eq:ndm_xt-1}
\end{equation}
with $\alpha_{t} = 1- \beta_{t}$ and $\Bar{\alpha}_{t} = \prod_{i=0}^{t}\alpha_{i}$, illustrated in \autoref{fig:ddnm}.

The term $\mathbf{z}_{t-1}$ represents a noised version of $\hat{\mathbf{z}}_{0|t}$. 
This noise effectively mitigates the dissonance between the range-space contents, represented by $\mathbf{A}^{\dagger}\mathbf{x}$, and the null-space contents, denoted by $(\mathbf{I} - \mathbf{A^{\dagger}}\mathbf{A})\mathbf{z}_{0|t}$.
The authors of DDNM show additionally that $\hat{\mathbf{z}}_{0|t}$ conforms to consistency.

The last step involves defining $\mathbf{A}$ and $\mathbf{A^{\dagger}}$, the construction of which is contingent on the restoration task at hand.
For instance, in SR tasks involving scaling by a factor of $n$, $\mathbf{A}$ can be defined as a $1 \times n^2$ matrix, representative of an average-pooling operator. 
The average-pooling operator, denoted as $\begin{bmatrix}\frac{1}{n^{2}} & ... & \frac{1}{n^{2}}\end{bmatrix}$, functions to average each patch into a singular value.
Similarly, we can construct its pseudo-inverse as $\mathbf{A}^{\dagger}\in\mathbb{R}^{n^2\times 1}=\begin{bmatrix}1 & ... & 1\end{bmatrix}^{\top}$.
The original work provides further examples of tasks (such as colorization, inpainting, and restoration), illustrating how these methods are applied. 
In addition, it describes how compound operations consisting of numerous sub-operations function in these contexts.
In their research, the authors also introduced DDNM$^+$ to support the restoration of noisy images. 
They utilized a technique analogous to the "back and forward" strategy implemented in RePaint \cite{lugmayr2022repaint}. 
This approach was leveraged to enhance the quality further.

Given this approach's novelty, only a handful of subsequent studies extend and build upon it, such as the work presented in CDPMSR \cite{niu2023cdpmsr}.
This research direction promises exciting possibilities, although it calls for further investigation.
For example, it should be noted that the DDNM approach introduces additional computational expenses compared to the task-specific training carried out using DDPMs. 
Moreover, the degradation operator $\mathbf{A}$ is set manually, which can be challenging for certain tasks.
Another potential drawback is the assumption that $\mathbf{A}$ functions as a linear degradation operator, which may not always hold true and thus could limit the model's effectiveness in certain scenarios.

\subsection{Posterior Estimation}

Most projection-based methods typically address the noiseless inverse problem. 
However, this assumption can weaken data consistency because the projection process can deviate the sample path from the data manifold \cite{li2023diffusion}.
To address this and enhance data consistency, some recent works \cite{chung2022diffusion, chung2022improving, song2022pseudoinverse} take a different approach by aiming to estimate the posterior distribution using the Bayes theorem:
\begin{equation}
    p(\mathbf{z}_t \mid \mathbf{x}) = \frac{p(\mathbf{x} \mid \mathbf{z}_t) \cdot p(\mathbf{z}_t)}{p(\mathbf{x})},
\end{equation}
This Bayesian approach provides a more robust and probabilistic framework for solving inverse problems, ultimately improving results in various image processing tasks. 
It results in the corresponding score function:
\begin{equation}
    \nabla_{\mathbf{z}_t} \log{p_t(\mathbf{z}_t \mid \mathbf{x})} = \nabla_{\mathbf{z}_t} \log{p_t(\mathbf{x} \mid \mathbf{z}_t)} + s_\theta(\mathbf{x}, t),
\end{equation}
where $s_\theta(\mathbf{x}, t)$ is extracted from a pre-trained model while $p_t(\mathbf{x} | \mathbf{z}_t)$ is intractable.
Thus, the goal is precisely estimating $p_t(\mathbf{x} | \mathbf{z}_t)$.
MCG \cite{chung2022improving} and DPS \cite{chung2022diffusion} approximate the posterior $p_t(\mathbf{x} | \mathbf{z}_t)$ with $p_t(\mathbf{x} | \mathbf{\hat{z}}_0(\mathbf{z}_t))$, where $\mathbf{\hat{z}}_0(\mathbf{z}_t)$ is the expectation given $\mathbf{z}_t$ as $\mathbf{\hat{z}}_0(\mathbf{z}_t) = \mathbb{E} \left[ \mathbf{z}_0 | \mathbf{z}_t\right]$ according to Tweedie's formula \cite{chung2022diffusion}.
While MCG also relies on projection, which can be harmful to data consistency, DPS discards the projection step and estimates the posterior as:
\begin{align}
    \nabla_{\mathbf{z}_t} \log{p_t(\mathbf{x} \mid \mathbf{z}_t)} &\approx \nabla_{\mathbf{z}_t} \log p(\mathbf{x} \mid \mathbf{\hat{z}}_0(\mathbf{z}_t)) \\
    & \approx -\frac{1}{\sigma^2} \nabla_{\mathbf{z}_t} \| \mathbf{x} - H(\mathbf{\hat{z}}_0(\mathbf{z}_t)) \|^2_2, \nonumber
\end{align}
where $H$ is a forward measurement operator.
A further expansion of this formula to the unified form for the linear, non-linear, differentiable inverse problem with Moore Penrose pseudoinverse can be found in IIGDM \cite{song2022pseudoinverse}.

A different approach to estimate $p_t(\mathbf{x} | \mathbf{z}_t)$ is demonstrated by GDP \cite{fei2023generative}.
The authors noted that a higher conditional probability of $p_t(\mathbf{x} | \mathbf{z}_t)$ correlates with a smaller distance between the application of the degradation model $\mathcal{D}(\mathbf{z}_t)$ and $\mathbf{x}$.
Thus, they propose a heuristic approximation:
\begin{equation}
    p_t(\mathbf{x} | \mathbf{z}_t) \approx \frac{1}{Z} \exp(-\left[ s\mathcal{L}(\mathcal{D}(\mathbf{z}_t), \mathbf{x})\right]) + \lambda \mathcal{Q} (\mathbf{z}_t),
\end{equation}
where $\mathcal{L}$ and $\mathcal{Q}$ denote a distance and quality metric, respectively. 
The term $Z$ is for normalization, and $s$ is a scaling factor controlling the guidance weight.
However, due to varying noise levels between $\mathbf{z}_t$ and $\mathbf{x}$, precisely defining the distance metric $\mathcal{L}$ can be challenging. 
To overcome this challenge, GDP substitutes $\mathbf{z}_t$ with its clean estimation $\mathbf{\hat{z}}_0$ in the distance calculation, providing a pragmatic solution to the noise discrepancy issue.

\section{Domain-Specific Applications}
\label{sec:aplications}
SR3 \cite{sr3-google} produces photo-realistic and perceptually state-of-the-art images on faces and natural images but may not be suitable for other tasks like remote sensing. 
Some models are more suited to certain tasks as they tackle issues specific to the domain \cite{lin2024adaptive}. 
This section highlights the applications of DMs to domain-specific SR tasks: Medical imaging, special cases of face SR (Blind Face Restoration and Atmospheric Turbulences), and remote sensing.

\subsection{Medical Imaging}
Magnetic Resonance Imaging (MRI) scans are widely used to aid patient diagnosis but can often be of low quality and corrupted with noise. 
Chung \textit{et al.} \cite{chung2022mr} propose a combined denoising and SR network referred to as R$2$D$2+$ (Regularized Reverse Diffusion Denoiser $+$ SR). 
They perform denoising of the MRI scans, followed by an SR module. 
Inspired by CCDF (i.e., a zero-shot method) from Chung \textit{et al.} \cite{chung2022ccdf}, they start their backward diffusion from an initial noisy image instead of pure Gaussian noise. 
The reverse SDE is solved using a non-parametric, eigenvalue-based method. 
In addition, they restrict the stochasticity of the DMs through low-frequency regularization. 
Particularly, they maintain low-frequency information while correcting the high-frequency ones to produce sharp and super-resolved MRI scans.  
Mao \textit{et al.} \cite{mao2023discdiff} addresses the lack of diffusion-based multi-contrast MRI SR methods. 
They propose a Disentangled Conditional Diffusion model (DisC-Diff) to leverage a multi-conditional fusion strategy based on representation disentanglement, enabling high-quality HR image sampling. 
Specifically, they employ a disentangled U-Net with multiple encoders to extract latent representations and use a novel joint disentanglement and Charbonnier loss function to learn representations across MRI contrasts. 
They also implement curriculum learning and improve their MRI model for varying anatomical complexity by gradually increasing the difficulty of training images.
An improvement of DisC-Diff by combining the DM with a transformer was introduced by Li \textit{et al.} with DiffMSR \cite{li2024rethinking}.

\subsection{Blind Face Restoration}
Most previously discussed SR methods are founded on a fixed degradation process during training, such as bicubic downsampling. 
However, when applied practically, these assumptions frequently diverge from the actual degradation process and yield subpar results. Additionally, datasets with pairs of clean and real-world distorted images are usually unavailable.
This issue is particularly researched in face SR, termed Blind Face Restoration (BFR), where datasets typically contain supervised samples $(\mathbf{x}, \mathbf{y})$ with unknown degradation.

A solution to BFR was proposed by Yue \textit{et al.} with DifFace \cite{yue2022difface} that leverages the rich generative priors of pre-trained DMs with parameters $\theta$, which were trained to approximate $p_\theta(\mathbf{z}_t | \mathbf{z}_{t-1})$. 
In contrast to existing methods that learn direct mappings from $\mathbf{x}$ to $\mathbf{y}$ under several constraints \cite{wang2021towards, yang2021gan}, DifFace circumvents this by generating a diffused version $\mathbf{z}_N$ of the desired HR image $\mathbf{y}$ with $N < T$.
They predict the starting point, the posterior $q(\mathbf{z}_N | \mathbf{x})$ via a transition distribution $p(\mathbf{z}_N | \mathbf{x})$.
The transition distribution is formulated like the regular diffusion process, a Gaussian distribution, but uses an initial predictor $\varphi(\mathbf{x})$ to generate the mean, named diffused estimator.
As their model borrows the reverse Markov chain from a pre-trained DM, DifFace requires no full retraining for new and unknown degradations, unlike SR3. 

A concurrent and better performing approach is DiffBFR \cite{qiu2023diffbfr} that adopts a two-step approach to BFR: A Identity Restoration Module (IRM), which employs two conditional DDPMs, and a Texture Enhancement Module (TEM), which employs an unconditional DDPM.
In the first step within the IRM, a conditional DDPM enriches facial details at a low-resolution space same as $\mathbf{x}$.
The downsampled version of $\mathbf{y}$ gives the target objective.
Next, it resizes the output to the desired spatial size of $\mathbf{y}$ and applies another conditional DDPM to approximate the HR image $\mathbf{y}$. 
To ensure minimal deviation from the actual image, DiffBFR employs a novel truncated sampling method, which begins denoising at intermediate steps.
The TEM further enhances realism through image texture and sharpened facial details. 
It imposes a diffuse-base facial prior with an unconditional DM trained on HR images and a backward diffusion starting from pure noise. 
However, it has more parameters than SR3 and requires optimization to accelerate sampling.   

Another method is DR$2$E\cite{wang2023dr2}, which employs two stages: degradation removal and enhancement modules. 
For degradation removal, they use a pre-trained face SR DDPM to remove degradations from an LR image with severe and unknown degradations. 
In particular, they diffuse the degraded image $\mathbf{x}$ in $T$ time steps to obtain $\mathbf{x}_T = \mathbf{z}_T$. 
Then, they use $\mathbf{x}_t$ to guide the backward diffusion such that the low-frequency part of $\mathbf{z}_t$ is replaced with that of $\mathbf{x}_t$, which is close in distribution. 
Theoretically, it produces visually clean intermediate results that are degradation-invariant. 
In the second stage, the enhancement module $p_\theta(y \mid \mathbf{z}_0)$, an arbitrary backbone CNN trained to map LR images to HR using a simple L$2$ loss, predicts the final output. 
DR$2$E can be slower than existing diffusion-based SR models for images with slight degradations and can even remove details from the input.

\subsection{Atmospheric Turbulence in Face SR}
Atmospheric Turbulence (AT) results from atmospheric conditions fluctuations, leading to images' perceptual degradation through geometric distortions, spatially variant blur, and noise. 
These alterations negatively impact downstream vision tasks, such as tracking or detection.
Wang \textit{et al.} \cite{wang2023atmospheric} introduced a variational inference framework known as AT-VarDiff, which aims to correct AT in generic scenes. 
The distinctive feature of this approach is its reliance on a conditioning signal derived from latent task-specific prior information extracted from the input image to guide the DM.
Nair \textit{et al.} \cite{nair2023ddpm} put forth another technique to restore facial images impaired by AT using SR. 
The method transfers class prior information from an SR model trained on clean facial data to a model designed to counteract turbulence degradation via knowledge distillation. 
The final model operates within the realistic faces manifold, which allows it to generate realistic face outputs even under substantial distortions. 
During inference, the process begins with noise- and turbulence-degraded images to ensure that the restored images closely resemble the distorted ones.

\subsection{Remote Sensing}
Remote Sensing Super-Resolution (RSSR) addresses the HR reconstruction from one or more LR images to aid object detection and semantic segmentation tasks for satellite imagery. 
RSSR is limited by the absence of small targets with complex granularity in the HR images \cite{xiao2023ediffsr}. 
To produce finer details and texture, Liu \textit{et al.} \cite{rs14194834} present DMs with a Detail Complement mechanism (DMDC). 
They train their model similar to SR3 \cite{sr3-google} and perform a detailed supplement task. 
To generate high-frequency information, they randomly mask several parts of the images to mimic dense objects. 
The SR images recover the occluded patches as the model learns small-grained information. 
Additionally, they introduce a novel pixel constraint loss to limit the diversity of DMDC and improve overall accuracy. 
Ali \textit{et al.} \cite{rs15092346} design a new architecture for RS images that integrates Vision Transformers (ViT) with DMs as a Two-stage approach for Enhancement and Super-Resolution (TESR). 
In the first stage (SR stage), the SwinIR \cite{liang2021swinir} model is used for RSSR. 
In the second stage (enhancement stage), the noisy images are enhanced by employing DMs to reconstruct the finer details.    
Xu \textit{et al.} \cite{xu2023dualdiffusion} propose a blind SR framework based on Dual conditioning DDPMs for SR (DDSR). A kernel predictor conditioned on LR image encodings estimates the degradation kernel in the first stage. This is followed by an SR module consisting of a conditional DDPM in a U-Net with the predicted kernel and the LR encodings as guidance. An RRDB encoder extracts the encodings from LR images.
Recently, Khanna \textit{et al.} introduced DiffusionSat \cite{khanna2023diffusionsat}, which uses a LDM for RSSR and incorporates additional remote sensing conditioning information (e.g., longitude, latitude, cloud cover, etc.).

\section{Discussion and Future Work}
\label{sec:discussion}
Though relatively new, DMs are quickly becoming a promising research area, especially in image SR. 
There are several avenues of ongoing research in this field, aiming to enhance the efficiency of DMs, accelerate computation speeds, and minimize memory footprint, all while generating high-quality, high-fidelity images.
This section introduces common problems of DMs for image SR and examines noteworthy research avenues for DMs specific to image SR.

\subsection{Color Shifting}
Often, the most practical advancements come from a solid theoretical understanding. 
As discussed in \autoref{sec:colorShift}, due to the substantial computational demands, DMs may occasionally exhibit color shifts when constrained by hardware limitations that demand smaller batch sizes or shorter training periods \cite{choi2022perception}.
While well-defined diffusion methods \cite{moser2023yoda} or color normalization \cite{wang2023exploiting} might mitigate this problem, a theoretical understanding of why it is emerging is necessary.

\subsection{Computational Costs}
In a study conducted by Ganguli \textit{et al.}, it was observed that the computing power needed for large-scale AI experiments has surged by over 300,000 times in the last decade \cite{ganguli2022predictability}. 
Regrettably, this increase in resource intensity has been accompanied by a sharp decline in the share of these results originating from academic circles. 
DMs are not immune to this issue; their computational demands add to the expanding gap between industry and academia. 
Therefore, there is a pressing need to reduce computational costs and memory footprints for practical applicability and research.
One strategy to alleviate computational demands is to examine smaller spatial-sized domains, as discussed in \autoref{sec:altDom}. 
Examples of such approaches include LDMs \cite{rombach-latent-diffusion-models, esser2021taming} and wavelet-based models \cite{moser2023waving, guth2022wavelet}. 
However, the capability of LDMs to reconstruct data with high precision and fine-grained accuracy, as required in image SR, remains to be questioned. 
Therefore, further advancements in these methods are critically needed.
On the other hand, wavelet-based models do not present a bottleneck regarding information preservation. 
This advantage suggests that they should be the subject of more intensive exploration.

\subsection{Efficient Sampling}
A benefit of DMs is the possibility of decoupling training and inference schedules \cite{chen2020wavegrad}. 
This allows for substantial enhancements in curtailing the time required for inference in practical applications, providing a significant efficiency edge in real-world scenarios. 
While reducing the number of steps taken during inference is relatively simple, a systematic method for determining inference schedules has yet to be developed \cite{cheng2023sampler}. 
As outlined in \autoref{sec:effSamp}, this research direction represents a promising avenue. 
We explored training-based sampling methods for SR with AddSR \cite{xie2024addsr} and YONOS-SR \cite{noroozi2024you} but also introduced efficient DMs that need fewer sampling steps, like ResShift \cite{yue2023resshift} and DiffIR \cite{xia2023diffir}.
An alternative is given by methods that use different corruption spaces, as discussed in \autoref{sec:corrupt}. 
Unlike sampling from pure Gaussian noise, notable works such as Luo \textit{et al.} \cite{luo2023image}, I$^2$SB \cite{liu20232}, Come-Closer-Diffuse-Faster \cite{chung2022ccdf}, or Cold Diffusion \cite{bansal2022cold} define a process from the LR to the HR image directly. 
Additional techniques for decreasing computation time, such as knowledge distillation, alternative noise schedulers, or truncated diffusion, demand further investigation concerning image SR \cite{xiao2021tackling, chen2023importance, distillation-guided-diffusion, knowledge-distillation}.

\subsection{Corruption Spaces}
New approaches for corruption spaces allow a more direct approach for upsampling images from LR to HR.
The significance of exploring different corruption spaces lies in addressing the inherent limitations and assumptions embedded within current DM frameworks, e.g., diversity and blurriness added during the forward diffusion process. 
The adaptability and efficiency demonstrated by novel approaches like InDI or I$^2$SB, especially in handling diverse and complex corruption patterns, spotlight the urgent need for future research.

\subsection{Comparability}
Comparing DMs in SR is complex because of the varied datasets used in different studies. 
They vary in resolution, content diversity, color distribution, and noise levels, all of which significantly influence model performance. 
A model may perform well with one dataset but poorly with another, complicating the assessment of its overall effectiveness.
Establishing a standard benchmark with diverse, representative datasets and uniform evaluation metrics is essential for comparability. 
This approach would help identify models that consistently perform well across different conditions and tasks, thereby promoting faster progress in the field.
Furthermore, evaluating the quality of SR images from generative models is still problematic. 
Although DMs often produce more photorealistic images, they typically score lower on standard metrics like PSNR and SSIM \cite{10041995}. 
However, these models tend to receive more favorable assessments from human evaluators \cite{sr3-google}.
LPIPS \cite{zhang2018unreasonable} performs better reflecting this perception, but the domain of image SR has to adapt to more diverse metrics, such as predictors that reflect human ratings directly \cite{liu2017rankiqa, ma2017dipiq}.
For instance, datasets with subjective ratings, like TID2013 \cite{ponomarenko2015image}, and neural networks, such as DeepQA \cite{kim2017deep} or NIMA \cite{talebi2018nima}, can be employed to predict human-like scoring of images and should be further explored.

\subsection{Image Manipulation}
Image manipulation can be particularly useful in multi-image SR for generating HR images that blend characteristics from multiple sources, potentially improving the quality and diversity of the output (e.g., satellite imagery for SR predictions with flexible daylights).
SRDiff \cite{srdiff} proposed two potential extensions: content fusion and latent space interpolation. 
Content fusion involves the combination of content from two source images. 
For instance, they replace the eyes in one source image with the face from another image before conducting diffusion in the image space like CutMix \cite{yun2019cutmix}. 
The backward diffusion successfully creates a smooth transition between both images.
In the latent space interpolation model, the latent space of two SR predictions is linearly interpolated to generate a new image. 
While these extensions have yielded remarkable results, unlike other generative models such as VAEs or GANs, DMs have been found to offer less proficient latent representations \cite{jing2022subspace}.
Therefore, recent and ongoing research into the manipulation of latent representations in DMs is both in its early stages and greatly needed \cite{kwon2023diffusion, wu2023uncovering, kim2022diffusionclip}.

\subsection{Cascaded Image Generation}
Saharia \textit{et al.} \cite{sr3-google} presented cascaded image SR, in which multiple DDPMs are chained across different scales. 
This strategy was applied to unconditional and class-conditional generation, cascading a model synthesizing $64 \times 64$ images with SR3 models generating $1024 \times 1024$ unconditional faces and $256 \times 256$ class-conditional natural images. 
The cascading approach allows several simpler models to be trained simultaneously, improving computational efficiency due to faster training times and reduced parameter counts.
Furthermore, they implemented cascading for inference, using more refinement steps at lower and fewer steps at higher resolutions. 
They found this more efficient than generating SR images directly. 
Even though their approach underperforms compared to BigGAN \cite{biggan} concerning cascaded generation, it still represents an exciting research opportunity.

\section{Conclusion}
\label{sec:conclusion}
Diffusion Models (DMs) revolutionized image Super-Resolution (SR) by enhancing both technical image quality and human perceptual preferences.
While traditional SR often focuses solely on pixel-level accuracy, DMs can generate HR images that are aesthetically pleasing and realistic.
Unlike previous generative models, they do not suffer typical convergence issues.
This survey explored the progress and diverse methods that have propelled DMs to the forefront of SR.
Potential use cases, as discussed in our applications section, extend far beyond what was previously imagined.
We introduced their foundational principles and compared them to other generative models. 
We explored conditioning strategies, from LR image guidance to text embeddings. 
Zero-shot SR, a particularly intriguing paradigm, was also a subject, as well as corruption spaces and image SR-specific topics like color shifting and architectural designs.
In conclusion, the survey provides a comprehensive guide to the current landscape and valuable insights into trends, challenges, and future directions.
As we continue to explore and refine these models, the future of image SR looks more promising than ever. 
%

\ifCLASSOPTIONcompsoc
  \section*{Acknowledgments}
\else
  \section*{Acknowledgment}
\fi
This work was supported by the BMBF project XAINES (Grant 01IW20005) and SustainML (Horizon Europe grant agreement No 101070408).

\ifCLASSOPTIONcaptionsoff
  \newpage
\fi

\bibliographystyle{IEEEtran}
\bibliography{references}

\begin{thebibliography}{100}
\providecommand{\url}[1]{#1}
\csname url@samestyle\endcsname
\providecommand{\newblock}{\relax}
\providecommand{\bibinfo}[2]{#2}
\providecommand{\BIBentrySTDinterwordspacing}{\spaceskip=0pt\relax}
\providecommand{\BIBentryALTinterwordstretchfactor}{4}
\providecommand{\BIBentryALTinterwordspacing}{\spaceskip=\fontdimen2\font plus
\BIBentryALTinterwordstretchfactor\fontdimen3\font minus \fontdimen4\font\relax}
\providecommand{\BIBforeignlanguage}[2]{{%
\expandafter\ifx\csname l@#1\endcsname\relax
\typeout{** WARNING: IEEEtran.bst: No hyphenation pattern has been}%
\typeout{** loaded for the language `#1'. Using the pattern for}%
\typeout{** the default language instead.}%
\else
\language=\csname l@#1\endcsname
\fi
#2}}
\providecommand{\BIBdecl}{\relax}
\BIBdecl

\bibitem{sun2020learned}
W.~Sun and Z.~Chen, ``Learned image downscaling for upscaling using content adaptive resampler,'' \emph{IEEE TIP}, vol.~29, 2020.

\bibitem{valsesia2021permutation}
D.~Valsesia and E.~Magli, ``Permutation invariance and uncertainty in multitemporal image super-resolution,'' \emph{IEEE Transactions on Geoscience and Remote Sensing}, vol.~60, 2021.

\bibitem{bashir2021comprehensive}
S.~M.~A. Bashir, Y.~Wang, M.~Khan, and Y.~Niu, ``A comprehensive review of deep learning-based single image super-resolution,'' \emph{PeerJ Computer Science}, vol.~7, 2021.

\bibitem{lee2022autoregressive}
D.~Lee, C.~Kim, S.~Kim, M.~Cho, and W.-S. Han, ``Autoregressive image generation using residual quantization,'' in \emph{CVPR}, 2022.

\bibitem{esser2021taming}
P.~Esser, R.~Rombach, and B.~Ommer, ``Taming transformers for high-resolution image synthesis,'' in \emph{CVPR}, 2021.

\bibitem{guo2022lar}
B.~Guo, X.~Zhang, H.~Wu, Y.~Wang, Y.~Zhang, and Y.-F. Wang, ``Lar-sr: A local autoregressive model for image super-resolution,'' in \emph{CVPR}, 2022.

\bibitem{frolov2021adversarial}
S.~Frolov, T.~Hinz, F.~Raue, J.~Hees, and A.~Dengel, ``Adversarial text-to-image synthesis: A review,'' \emph{Neural Networks}, vol. 144, 2021.

\bibitem{ho2020denoising}
J.~Ho, A.~Jain, and P.~Abbeel, ``Denoising diffusion probabilistic models,'' \emph{NeurIPS}, vol.~33, 2020.

\bibitem{goodfellow2020generative}
I.~Goodfellow, J.~Pouget-Abadie, M.~Mirza, B.~Xu, D.~Warde-Farley, S.~Ozair, A.~Courville, and Y.~Bengio, ``Generative adversarial networks,'' \emph{Communications of the ACM}, vol.~63, no.~11, 2020.

\bibitem{song2019generative}
Y.~Song and S.~Ermon, ``Generative modeling by estimating gradients of the data distribution,'' \emph{NeurIPS}, vol.~32, 2019.

\bibitem{song2020score}
Y.~Song, J.~Sohl-Dickstein, D.~P. Kingma, A.~Kumar, S.~Ermon, and B.~Poole, ``Score-based generative modeling through stochastic differential equations,'' \emph{arXiv:2011.13456}, 2020.

\bibitem{dhariwal-diffusion-models-beat-gans}
P.~Dhariwal and A.~Nichol, ``Diffusion models beat gans on image synthesis,'' \emph{NeurIPS}, vol.~34, 2021.

\bibitem{rombach-latent-diffusion-models}
R.~Rombach, A.~Blattmann, D.~Lorenz, P.~Esser, and B.~Ommer, ``High-resolution image synthesis with latent diffusion models,'' in \emph{CVPR}, 2022.

\bibitem{ramesh2022hierarchical}
A.~Ramesh, P.~Dhariwal, A.~Nichol, C.~Chu, and M.~Chen, ``Hierarchical text-conditional image generation with clip latents,'' \emph{arXiv:2204.06125}, 2022.

\bibitem{sr3-google}
C.~Saharia, J.~Ho, W.~Chan, T.~Salimans, D.~J. Fleet, and M.~Norouzi, ``Image super-resolution via iterative refinement,'' \emph{IEEE TPAMI}, vol.~45, no.~4, 2023.

\bibitem{10041995}
B.~B. Moser, F.~Raue, S.~Frolov, S.~Palacio, J.~Hees, and A.~Dengel, ``Hitchhiker's guide to super-resolution: Introduction and recent advances,'' \emph{IEEE TPAMI}, 2023.

\bibitem{li2023diffusion}
X.~Li, Y.~Ren, X.~Jin, C.~Lan, X.~Wang, W.~Zeng, X.~Wang, and Z.~Chen, ``Diffusion models for image restoration and enhancement--a comprehensive survey,'' \emph{arXiv:2308.09388}, 2023.

\bibitem{liu2022blind}
A.~Liu, Y.~Liu, J.~Gu, Y.~Qiao, and C.~Dong, ``Blind image super-resolution: A survey and beyond,'' \emph{IEEE TPAMI}, vol.~45, no.~5, 2022.

\bibitem{anwar2020densely}
S.~Anwar and N.~Barnes, ``Densely residual laplacian super-resolution,'' \emph{IEEE TPAMI}, 2020.

\bibitem{MATLAB:2017b}
\emph{{MATLAB}}, The Mathworks, Inc., Natick, Massachusetts, 2017.

\bibitem{div2k}
E.~Agustsson and R.~Timofte, ``Ntire 2017 challenge on single image super-resolution: Dataset and study,'' in \emph{CVPRW}, July 2017.

\bibitem{bevilacqua2012low}
M.~Bevilacqua, A.~Roumy, C.~Guillemot, and M.~L. Alberi-Morel, ``Low-complexity single-image super-resolution based on nonnegative neighbor embedding,'' 2012.

\bibitem{zeyde2010single}
R.~Zeyde, M.~Elad, and M.~Protter, ``On single image scale-up using sparse-representations,'' in \emph{International conference on curves and surfaces}.\hskip 1em plus 0.5em minus 0.4em\relax Springer, 2010.

\bibitem{martin2001database}
D.~Martin, C.~Fowlkes, D.~Tal, and J.~Malik, ``A database of human segmented natural images and its application to evaluating segmentation algorithms and measuring ecological statistics,'' in \emph{ICCV}, vol.~2.\hskip 1em plus 0.5em minus 0.4em\relax IEEE, 2001.

\bibitem{huang2015single}
J.-B. Huang, A.~Singh, and N.~Ahuja, ``Single image super-resolution from transformed self-exemplars,'' in \emph{CVPR}, 2015.

\bibitem{matsui2017sketch}
Y.~Matsui, K.~Ito, Y.~Aramaki, A.~Fujimoto, T.~Ogawa, T.~Yamasaki, and K.~Aizawa, ``Sketch-based manga retrieval using manga109 dataset,'' \emph{Multimedia Tools and Applications}, vol.~76, no.~20, 2017.

\bibitem{flickr2k}
E.~Agustsson and R.~Timofte, ``Ntire 2017 challenge on single image super-resolution: Dataset and study,'' in \emph{CVPRW}, 2017.

\bibitem{ffhq}
T.~Karras, S.~Laine, and T.~Aila, ``A style-based generator architecture for generative adversarial networks,'' in \emph{CVPR}, 2019.

\bibitem{celeba-hq}
T.~Karras, T.~Aila, S.~Laine, and J.~Lehtinen, ``Progressive growing of gans for improved quality, stability, and variation,'' \emph{arXiv:1710.10196}, 2017.

\bibitem{imagenet}
J.~Deng, W.~Dong, R.~Socher, L.-J. Li, K.~Li, and L.~Fei-Fei, ``Imagenet: A large-scale hierarchical image database,'' in \emph{CVPR}, 2009.

\bibitem{voc-2012}
M.~Everingham, L.~Van~Gool, C.~K.~I. Williams, J.~Winn, and A.~Zisserman, ``The {PASCAL} voc2012 {R}esults,'' http://www.pascal-network.org/challenges/VOC/voc2012/workshop/index.html, 2012.

\bibitem{stat-based-1}
K.~I. Kim and Y.~Kwon, ``Single-image super-resolution using sparse regression and natural image prior,'' \emph{IEEE TPAMI}, vol.~32, no.~6, 2010.

\bibitem{edge-based-1}
G.~Freedman and R.~Fattal, ``Image and video upscaling from local self-examples,'' \emph{ACM Trans. Graph.}, vol.~30, no.~2, apr 2011.

\bibitem{edge-based-2}
J.~Sun, Z.~Xu, and H.-Y. Shum, ``Image super-resolution using gradient profile prior,'' in \emph{CVPR}, 2008.

\bibitem{patch-based-1}
H.~Chang, D.-Y. Yeung, and Y.~Xiong, ``Super-resolution through neighbor embedding,'' in \emph{Proceedings of the 2004 IEEE Computer Society Conference on Computer Vision and Pattern Recognition, 2004. CVPR 2004.}, vol.~1.\hskip 1em plus 0.5em minus 0.4em\relax IEEE, 2004.

\bibitem{patch-based-2}
W.~Freeman, T.~Jones, and E.~Pasztor, ``Example-based super-resolution,'' \emph{IEEE Computer Graphics and Applications}, vol.~22, no.~2, 2002.

\bibitem{pred-based-1}
R.~Keys, ``Cubic convolution interpolation for digital image processing,'' \emph{IEEE Transactions on Acoustics, Speech, and Signal Processing}, vol.~29, no.~6, 1981.

\bibitem{pred-based-2}
M.~Irani and S.~Peleg, ``Improving resolution by image registration,'' \emph{CVGIP: Graphical Models and Image Processing}, vol.~53, no.~3, 1991.

\bibitem{sparse-based-1}
J.~Yang, J.~Wright, T.~S. Huang, and Y.~Ma, ``Image super-resolution via sparse representation,'' \emph{IEEE TIP}, vol.~19, no.~11, 2010.

\bibitem{srcnn}
C.~Dong, C.~C. Loy, K.~He, and X.~Tang, ``Image super-resolution using deep convolutional networks,'' \emph{IEEE TPAMI}, vol.~38, no.~2, 2015.

\bibitem{fsr-cnn}
C.~Dong, C.~C. Loy, and X.~Tang, ``Accelerating the super-resolution convolutional neural network,'' in \emph{ECCV}.\hskip 1em plus 0.5em minus 0.4em\relax Springer, 2016.

\bibitem{espcnn}
W.~Shi, J.~Caballero, F.~Husz{\'a}r, J.~Totz, A.~P. Aitken, R.~Bishop, D.~Rueckert, and Z.~Wang, ``Real-time single image and video super-resolution using an efficient sub-pixel convolutional neural network,'' in \emph{CVPR}, 2016.

\bibitem{srresnet}
C.~Ledig, L.~Theis, F.~Husz{\'a}r, J.~Caballero, A.~Cunningham, A.~Acosta, A.~Aitken, A.~Tejani, J.~Totz, Z.~Wang \emph{et~al.}, ``Photo-realistic single image super-resolution using a generative adversarial network,'' in \emph{CVPR}, 2017.

\bibitem{densenet}
G.~Huang, Z.~Liu, L.~Van Der~Maaten, and K.~Q. Weinberger, ``Densely connected convolutional networks,'' in \emph{CVPR}, 2017.

\bibitem{srdensenet}
T.~Tong, G.~Li, X.~Liu, and Q.~Gao, ``Image super-resolution using dense skip connections,'' in \emph{ICCV}, 2017.

\bibitem{drcn}
J.~Kim, J.~K. Lee, and K.~M. Lee, ``Deeply-recursive convolutional network for image super-resolution,'' in \emph{CVPR}, 2016.

\bibitem{drrn}
Y.~Tai, J.~Yang, and X.~Liu, ``Image super-resolution via deep recursive residual network,'' in \emph{CVPR}, 2017.

\bibitem{carn}
N.~Ahn, B.~Kang, and K.-A. Sohn, ``Fast, accurate, and lightweight super-resolution with cascading residual network,'' in \emph{ECCV}, 2018.

\bibitem{liang2021swinir}
J.~Liang, J.~Cao, G.~Sun, K.~Zhang, L.~Van~Gool, and R.~Timofte, ``Swinir: Image restoration using swin transformer,'' in \emph{CVPR}, 2021.

\bibitem{chen2023activating}
X.~Chen, X.~Wang, J.~Zhou, Y.~Qiao, and C.~Dong, ``Activating more pixels in image super-resolution transformer,'' in \emph{CVPR}, 2023.

\bibitem{hsu2024drct}
C.-C. Hsu, C.-M. Lee, and Y.-S. Chou, ``Drct: Saving image super-resolution away from information bottleneck,'' \emph{arXiv:2404.00722}, 2024.

\bibitem{srgan}
C.~Ledig, L.~Theis, F.~Husz{\'a}r, J.~Caballero, A.~Cunningham, A.~Acosta, A.~Aitken, A.~Tejani, J.~Totz, Z.~Wang \emph{et~al.}, ``Photo-realistic single image super-resolution using a generative adversarial network,'' in \emph{CVPR}, 2017.

\bibitem{esrgan}
X.~Wang, K.~Yu, S.~Wu, J.~Gu, Y.~Liu, C.~Dong, Y.~Qiao, and C.~Change~Loy, ``Esrgan: Enhanced super-resolution generative adversarial networks,'' 2018.

\bibitem{srflow}
A.~Lugmayr, M.~Danelljan, L.~Van~Gool, and R.~Timofte, ``Srflow: Learning the super-resolution space with normalizing flow,'' in \emph{ECCV}.\hskip 1em plus 0.5em minus 0.4em\relax Springer, 2020.

\bibitem{simonyan2014very}
K.~Simonyan and A.~Zisserman, ``Very deep convolutional networks for large-scale image recognition,'' \emph{arXiv:1409.1556}, 2014.

\bibitem{krizhevsky2017imagenet}
A.~Krizhevsky, I.~Sutskever, and G.~E. Hinton, ``Imagenet classification with deep convolutional neural networks,'' \emph{Communications of the ACM}, vol.~60, no.~6, 2017.

\bibitem{mittal2012no}
A.~Mittal, A.~K. Moorthy, and A.~C. Bovik, ``No-reference image quality assessment in the spatial domain,'' \emph{IEEE TIP}, vol.~21, no.~12, 2012.

\bibitem{mittal2012making}
A.~Mittal, R.~Soundararajan, and A.~C. Bovik, ``Making a “completely blind” image quality analyzer,'' \emph{IEEE Signal processing letters}, vol.~20, no.~3, 2012.

\bibitem{radford2021learning}
A.~Radford, J.~W. Kim, C.~Hallacy, A.~Ramesh, G.~Goh, S.~Agarwal, G.~Sastry, A.~Askell, P.~Mishkin, J.~Clark \emph{et~al.}, ``Learning transferable visual models from natural language supervision,'' in \emph{ICML}.\hskip 1em plus 0.5em minus 0.4em\relax PMLR, 2021.

\bibitem{wang2023exploring}
J.~Wang, K.~C. Chan, and C.~C. Loy, ``Exploring clip for assessing the look and feel of images,'' in \emph{AAAI}, vol.~37, no.~2, 2023.

\bibitem{ponomarenko2015image}
N.~Ponomarenko, L.~Jin, O.~Ieremeiev, V.~Lukin, K.~Egiazarian, J.~Astola, B.~Vozel, K.~Chehdi, M.~Carli, F.~Battisti \emph{et~al.}, ``Image database tid2013: Peculiarities, results and perspectives,'' \emph{Signal processing: Image communication}, vol.~30, 2015.

\bibitem{kim2017deep}
J.~Kim and S.~Lee, ``Deep learning of human visual sensitivity in image quality assessment framework,'' in \emph{CVPR}, 2017.

\bibitem{talebi2018nima}
H.~Talebi and P.~Milanfar, ``Nima: Neural image assessment,'' \emph{IEEE TIP}, vol.~27, no.~8, 2018.

\bibitem{ke2021musiq}
J.~Ke, Q.~Wang, Y.~Wang, P.~Milanfar, and F.~Yang, ``Musiq: Multi-scale image quality transformer,'' in \emph{ICCV}, 2021.

\bibitem{yang2023diffusion}
L.~Yang, Z.~Zhang, Y.~Song, S.~Hong, R.~Xu, Y.~Zhao, W.~Zhang, B.~Cui, and M.-H. Yang, ``Diffusion models: A comprehensive survey of methods and applications,'' \emph{ACM Computing Surveys}, vol.~56, no.~4, 2023.

\bibitem{sohl2015deep}
J.~Sohl-Dickstein, E.~Weiss, N.~Maheswaranathan, and S.~Ganguli, ``Deep unsupervised learning using nonequilibrium thermodynamics,'' in \emph{ICML}.\hskip 1em plus 0.5em minus 0.4em\relax PMLR, 2015.

\bibitem{parisi1981correlation}
G.~Parisi, ``Correlation functions and computer simulations,'' \emph{Nuclear Physics B}, vol. 180, no.~3, 1981.

\bibitem{vincent2011connection}
P.~Vincent, ``A connection between score matching and denoising autoencoders,'' \emph{Neural computation}, vol.~23, no.~7, 2011.

\bibitem{hyvarinen2005estimation}
A.~Hyv{\"a}rinen and P.~Dayan, ``Estimation of non-normalized statistical models by score matching.'' \emph{Journal of Machine Learning Research}, vol.~6, no.~4, 2005.

\bibitem{song2020sliced}
Y.~Song, S.~Garg, J.~Shi, and S.~Ermon, ``Sliced score matching: A scalable approach to density and score estimation,'' in \emph{Uncertainty in Artificial Intelligence}.\hskip 1em plus 0.5em minus 0.4em\relax PMLR, 2020.

\bibitem{anderson1982reverse}
B.~D. Anderson, ``Reverse-time diffusion equation models,'' \emph{Stochastic Processes and their Applications}, vol.~12, no.~3, 1982.

\bibitem{gans}
I.~Goodfellow, J.~Pouget-Abadie, M.~Mirza, B.~Xu, D.~Warde-Farley, S.~Ozair, A.~Courville, and Y.~Bengio, ``Generative adversarial nets,'' \emph{NeurIPS}, vol.~27, 2014.

\bibitem{vae}
D.~P. Kingma and M.~Welling, ``Auto-encoding variational bayes,'' \emph{arXiv:1312.6114}, 2013.

\bibitem{normalizing-flows}
D.~Rezende and S.~Mohamed, ``Variational inference with normalizing flows,'' in \emph{ICML}.\hskip 1em plus 0.5em minus 0.4em\relax PMLR, 2015.

\bibitem{diff-flow}
Q.~Zhang and Y.~Chen, ``Diffusion normalizing flow,'' in \emph{NeurIPS}, M.~Ranzato, A.~Beygelzimer, Y.~Dauphin, P.~Liang, and J.~W. Vaughan, Eds., vol.~34.\hskip 1em plus 0.5em minus 0.4em\relax Curran Associates, Inc., 2021.

\bibitem{papamakarios2021normalizing}
G.~Papamakarios, E.~Nalisnick, D.~J. Rezende, S.~Mohamed, and B.~Lakshminarayanan, ``Normalizing flows for probabilistic modeling and inference,'' \emph{The Journal of Machine Learning Research}, vol.~22, no.~1, 2021.

\bibitem{karras2022elucidating}
T.~Karras, M.~Aittala, T.~Aila, and S.~Laine, ``Elucidating the design space of diffusion-based generative models,'' \emph{NeurIPS}, vol.~35, 2022.

\bibitem{vgg-face}
``Oxford vggface implementation using keras functional framework v2+,'' \url{https://github.com/rcmalli/keras-vggface}.

\bibitem{srdiff}
H.~Li, Y.~Yang, M.~Chang, S.~Chen, H.~Feng, Z.~Xu, Q.~Li, and Y.~Chen, ``Srdiff: Single image super-resolution with diffusion probabilistic models,'' \emph{Neurocomputing}, vol. 479, 2022.

\bibitem{fast-sampling-ddpm-1}
D.~Watson, J.~Ho, M.~Norouzi, and W.~Chan, ``Learning to efficiently sample from diffusion probabilistic models,'' \emph{arXiv:2106.03802}, 2021.

\bibitem{fast-sampling-ddpm-2}
D.~Watson, W.~Chan, J.~Ho, and M.~Norouzi, ``Learning fast samplers for diffusion models by differentiating through sample quality,'' \emph{arXiv:2202.05830}, 2022.

\bibitem{truncated-diffusion-1}
Z.~Lyu, X.~Xu, C.~Yang, D.~Lin, and B.~Dai, ``Accelerating diffusion models via early stop of the diffusion process,'' \emph{arXiv:2205.12524}, 2022.

\bibitem{truncated-diffusion-2}
L.~Zhang, A.~Rao, and M.~Agrawala, ``Adding conditional control to text-to-image diffusion models,'' in \emph{CVPR}, 2023.

\bibitem{distillation-guided-diffusion}
C.~Meng, R.~Rombach, R.~Gao, D.~Kingma, S.~Ermon, J.~Ho, and T.~Salimans, ``On distillation of guided diffusion models,'' in \emph{CVPR}, 2023.

\bibitem{knowledge-distillation}
E.~Luhman and T.~Luhman, ``Knowledge distillation in iterative generative models for improved sampling speed,'' \emph{arXiv:2101.02388}, 2021.

\bibitem{progressive-distillation-diffusion}
T.~Salimans and J.~Ho, ``Progressive distillation for fast sampling of diffusion models,'' \emph{arXiv:2202.00512}, 2022.

\bibitem{Denoising-Diffusion-GANs}
Z.~Xiao, K.~Kreis, and A.~Vahdat, ``Tackling the generative learning trilemma with denoising diffusion gans,'' \emph{arXiv:2112.07804}, 2021.

\bibitem{xie2024addsr}
R.~Xie, Y.~Tai, K.~Zhang, Z.~Zhang, J.~Zhou, and J.~Yang, ``Addsr: Accelerating diffusion-based blind super-resolution with adversarial diffusion distillation,'' \emph{arXiv:2404.01717}, 2024.

\bibitem{noroozi2024you}
M.~Noroozi, I.~Hadji, B.~Martinez, A.~Bulat, and G.~Tzimiropoulos, ``You only need one step: Fast super-resolution with stable diffusion via scale distillation,'' \emph{arXiv:2401.17258}, 2024.

\bibitem{ddim}
J.~Song, C.~Meng, and S.~Ermon, ``Denoising diffusion implicit models,'' \emph{arXiv:2010.02502}, 2020.

\bibitem{score-based-fast-generation}
A.~Jolicoeur-Martineau, K.~Li, R.~Pich{\'e}-Taillefer, T.~Kachman, and I.~Mitliagkas, ``Gotta go fast when generating data with score-based models,'' \emph{arXiv:2105.14080}, 2021.

\bibitem{lu2022dpm}
C.~Lu, Y.~Zhou, F.~Bao, J.~Chen, C.~Li, and J.~Zhu, ``Dpm-solver: A fast ode solver for diffusion probabilistic model sampling in around 10 steps,'' \emph{Advances in Neural Information Processing Systems}, vol.~35, pp. 5775--5787, 2022.

\bibitem{bao2022analytic}
F.~Bao, C.~Li, J.~Zhu, and B.~Zhang, ``Analytic-dpm: an analytic estimate of the optimal reverse variance in diffusion probabilistic models,'' \emph{arXiv:2201.06503}, 2022.

\bibitem{lu2022dpm2}
C.~Lu, Y.~Zhou, F.~Bao, J.~Chen, C.~Li, and J.~Zhu, ``Dpm-solver++: Fast solver for guided sampling of diffusion probabilistic models,'' \emph{arXiv:2211.01095}, 2022.

\bibitem{zhao2024unipc}
W.~Zhao, L.~Bai, Y.~Rao, J.~Zhou, and J.~Lu, ``Unipc: A unified predictor-corrector framework for fast sampling of diffusion models,'' \emph{NeurIPS}, vol.~36, 2024.

\bibitem{ho2019compression}
J.~Ho, E.~Lohn, and P.~Abbeel, ``Compression with flows via local bits-back coding,'' \emph{NeurIPS}, vol.~32, 2019.

\bibitem{dai2017good}
Z.~Dai, Z.~Yang, F.~Yang, W.~W. Cohen, and R.~R. Salakhutdinov, ``Good semi-supervised learning that requires a bad gan,'' \emph{NeurIPS}, vol.~30, 2017.

\bibitem{max-likelihood-diffusion}
Y.~Song, C.~Durkan, I.~Murray, and S.~Ermon, ``Maximum likelihood training of score-based diffusion models,'' \emph{NeurIPS}, vol.~34, 2021.

\bibitem{variational-dm}
D.~Kingma, T.~Salimans, B.~Poole, and J.~Ho, ``Variational diffusion models,'' \emph{NeurIPS}, vol.~34, 2021.

\bibitem{improved-ddpm}
A.~Q. Nichol and P.~Dhariwal, ``Improved denoising diffusion probabilistic models,'' in \emph{ICML}.\hskip 1em plus 0.5em minus 0.4em\relax PMLR, 2021.

\bibitem{ho2022cascaded}
J.~Ho, C.~Saharia, W.~Chan, D.~J. Fleet, M.~Norouzi, and T.~Salimans, ``Cascaded diffusion models for high fidelity image generation.'' \emph{J. Mach. Learn. Res.}, vol.~23, no.~47, 2022.

\bibitem{biggan}
A.~Brock, J.~Donahue, and K.~Simonyan, ``Large scale gan training for high fidelity natural image synthesis,'' \emph{arXiv:1809.11096}, 2018.

\bibitem{ho2022classifier}
J.~Ho and T.~Salimans, ``Classifier-free diffusion guidance,'' \emph{arXiv:2207.12598}, 2022.

\bibitem{luo2022understanding}
C.~Luo, ``Understanding diffusion models: A unified perspective,'' \emph{arXiv:2208.11970}, 2022.

\bibitem{kim2024arbitrary}
J.~Kim and T.-K. Kim, ``Arbitrary-scale image generation and upsampling using latent diffusion model and implicit neural decoder,'' \emph{arXiv:2403.10255}, 2024.

\bibitem{score-based-ldm}
A.~Vahdat, K.~Kreis, and J.~Kautz, ``Score-based generative modeling in latent space,'' \emph{NeurIPS}, vol.~34, 2021.

\bibitem{wang2023exploiting}
J.~Wang, Z.~Yue, S.~Zhou, K.~C. Chan, and C.~C. Loy, ``Exploiting diffusion prior for real-world image super-resolution,'' \emph{arXiv:2305.07015}, 2023.

\bibitem{luo2023image}
Z.~Luo, F.~K. Gustafsson, Z.~Zhao, J.~Sj{\"o}lund, and T.~B. Sch{\"o}n, ``Image restoration with mean-reverting stochastic differential equations,'' \emph{arXiv:2301.11699}, 2023.

\bibitem{chen2022simple}
L.~Chen, X.~Chu, X.~Zhang, and J.~Sun, ``Simple baselines for image restoration,'' in \emph{ECCV}.\hskip 1em plus 0.5em minus 0.4em\relax Springer, 2022.

\bibitem{chen2023hierarchical}
Z.~Chen, Y.~Zhang, D.~Liu, B.~Xia, J.~Gu, L.~Kong, and X.~Yuan, ``Hierarchical integration diffusion model for realistic image deblurring,'' \emph{arXiv:2305.12966}, 2023.

\bibitem{10.1007/978-3-031-44210-0_19}
B.~B. Moser, S.~Frolov, F.~Raue, S.~Palacio, and A.~Dengel, ``Dwa: Differential wavelet amplifier for image super-resolution,'' in \emph{Artificial Neural Networks and Machine Learning -- ICANN 2023}, L.~Iliadis, A.~Papaleonidas, P.~Angelov, and C.~Jayne, Eds.\hskip 1em plus 0.5em minus 0.4em\relax Cham: Springer Nature Switzerland, 2023.

\bibitem{guo2017deep}
T.~Guo, H.~Seyed~Mousavi, T.~Huu~Vu, and V.~Monga, ``Deep wavelet prediction for image super-resolution,'' in \emph{CVPRW}, 2017.

\bibitem{moser2023waving}
B.~Moser, S.~Frolov, F.~Raue, S.~Palacio, and A.~Dengel, ``Waving goodbye to low-res: A diffusion-wavelet approach for image super-resolution,'' 2023.

\bibitem{huang2023wavedm}
Y.~Huang, J.~Huang, J.~Liu, Y.~Dong, J.~Lv, and S.~Chen, ``Wavedm: Wavelet-based diffusion models for image restoration,'' \emph{arXiv:2305.13819}, 2023.

\bibitem{guth2022wavelet}
F.~Guth, S.~Coste, V.~De~Bortoli, and S.~Mallat, ``Wavelet score-based generative modeling,'' \emph{NeurIPS}, vol.~35, 2022.

\bibitem{shang2023resdiff}
S.~Shang, Z.~Shan, G.~Liu, and J.~Zhang, ``Resdiff: Combining cnn and diffusion model for image super-resolution,'' \emph{arXiv:2303.08714}, 2023.

\bibitem{deblur-diff}
J.~Whang, M.~Delbracio, H.~Talebi, C.~Saharia, A.~G. Dimakis, and P.~Milanfar, ``Deblurring via stochastic refinement,'' in \emph{CVPR}, 2022.

\bibitem{yue2023resshift}
Z.~Yue, J.~Wang, and C.~C. Loy, ``Resshift: Efficient diffusion model for image super-resolution by residual shifting,'' 2023.

\bibitem{saharia2022photorealistic}
C.~Saharia, W.~Chan, S.~Saxena, L.~Li, J.~Whang, E.~L. Denton, K.~Ghasemipour, R.~Gontijo~Lopes, B.~Karagol~Ayan, T.~Salimans \emph{et~al.}, ``Photorealistic text-to-image diffusion models with deep language understanding,'' \emph{NeurIPS}, vol.~35, 2022.

\bibitem{choi2021ilvr}
J.~Choi, S.~Kim, Y.~Jeong, Y.~Gwon, and S.~Yoon, ``Ilvr: Conditioning method for denoising diffusion probabilistic models,'' 2021.

\bibitem{niu2023cdpmsr}
A.~Niu, K.~Zhang, T.~X. Pham, J.~Sun, Y.~Zhu, I.~S. Kweon, and Y.~Zhang, ``Cdpmsr: Conditional diffusion probabilistic models for single image super-resolution,'' 2023.

\bibitem{lim2017enhanced}
B.~Lim, S.~Son, H.~Kim, S.~Nah, and K.~Mu~Lee, ``Enhanced deep residual networks for single image super-resolution,'' in \emph{CVPRW}, 2017.

\bibitem{park2022flexible}
S.~H. Park, Y.~S. Moon, and N.~I. Cho, ``Flexible style image super-resolution using conditional objective,'' \emph{IEEE Access}, vol.~10, 2022.

\bibitem{zhang2019ranksrgan}
W.~Zhang, Y.~Liu, C.~Dong, and Y.~Qiao, ``Ranksrgan: Generative adversarial networks with ranker for image super-resolution,'' in \emph{CVPR}, 2019.

\bibitem{lugmayr2020srflow}
A.~Lugmayr, M.~Danelljan, L.~V. Gool, and R.~Timofte, ``Srflow: Learning the super-resolution space with normalizing flow,'' in \emph{ECCV}.\hskip 1em plus 0.5em minus 0.4em\relax Springer, 2020.

\bibitem{pulse}
S.~Menon, A.~Damian, S.~Hu, N.~Ravi, and C.~Rudin, ``Pulse: Self-supervised photo upsampling via latent space exploration of generative models,'' in \emph{CVPR}, 2020.

\bibitem{chen2018fsrnet}
Y.~Chen, Y.~Tai, X.~Liu, C.~Shen, and J.~Yang, ``Fsrnet: End-to-end learning face super-resolution with facial priors,'' in \emph{CVPR}, 2018.

\bibitem{pandey2022diffusevae}
K.~Pandey, A.~Mukherjee, P.~Rai, and A.~Kumar, ``Diffusevae: Efficient, controllable and high-fidelity generation from low-dimensional latents,'' \emph{arXiv:2201.00308}, 2022.

\bibitem{bi2024deedsr}
C.~Bi, X.~Luo, S.~Shen, M.~Zhang, H.~Yue, and J.~Yang, ``Deedsr: Towards real-world image super-resolution via degradation-aware stable diffusion,'' \emph{arXiv}, 2024.

\bibitem{zhou2022towards}
S.~Zhou, K.~Chan, C.~Li, and C.~C. Loy, ``Towards robust blind face restoration with codebook lookup transformer,'' \emph{NeurIPS}, vol.~35, 2022.

\bibitem{yang2023pixel}
T.~Yang, P.~Ren, X.~Xie, and L.~Zhang, ``Pixel-aware stable diffusion for realistic image super-resolution and personalized stylization,'' \emph{arXiv:2308.14469}, 2023.

\bibitem{wu2023seesr}
R.~Wu, T.~Yang, L.~Sun, Z.~Zhang, S.~Li, and L.~Zhang, ``Seesr: Towards semantics-aware real-world image super-resolution,'' \emph{arXiv:2311.16518}, 2023.

\bibitem{qu2024xpsr}
Y.~Qu, K.~Yuan, K.~Zhao, Q.~Xie, J.~Hao, M.~Sun, and C.~Zhou, ``Xpsr: Cross-modal priors for diffusion-based image super-resolution,'' \emph{arXiv:2403.05049}, 2024.

\bibitem{zhang2021designing}
K.~Zhang, J.~Liang, L.~Van~Gool, and R.~Timofte, ``Designing a practical degradation model for deep blind image super-resolution,'' in \emph{ICCV}, 2021.

\bibitem{wang2021real}
X.~Wang, L.~Xie, C.~Dong, and Y.~Shan, ``Real-esrgan: Training real-world blind super-resolution with pure synthetic data,'' in \emph{CVPR}, 2021.

\bibitem{liang2022details}
J.~Liang, H.~Zeng, and L.~Zhang, ``Details or artifacts: A locally discriminative learning approach to realistic image super-resolution,'' in \emph{CVPR}, 2022.

\bibitem{chen2022femasr}
C.~Chen, X.~Shi, Y.~Qin, X.~Li, X.~Han, T.~Yang, and S.~Guo, ``Real-world blind super-resolution via feature matching with implicit high-resolution priors,'' in \emph{Proceedings of the 30th ACM International Conference on Multimedia}, ser. MM '22.\hskip 1em plus 0.5em minus 0.4em\relax New York, NY, USA: Association for Computing Machinery, 2022.

\bibitem{daras2022soft}
G.~Daras, M.~Delbracio, H.~Talebi, A.~G. Dimakis, and P.~Milanfar, ``Soft diffusion: Score matching for general corruptions,'' \emph{arXiv:2209.05442}, 2022.

\bibitem{bansal2022cold}
A.~Bansal, E.~Borgnia, H.-M. Chu, J.~S. Li, H.~Kazemi, F.~Huang, M.~Goldblum, J.~Geiping, and T.~Goldstein, ``Cold diffusion: Inverting arbitrary image transforms without noise,'' \emph{arXiv:2208.09392}, 2022.

\bibitem{liu20232}
G.-H. Liu, A.~Vahdat, D.-A. Huang, E.~A. Theodorou, W.~Nie, and A.~Anandkumar, ``I $^2$ sb: Image-to-image schr\"{o}dinger bridge,'' \emph{arXiv:2302.05872}, 2023.

\bibitem{delbracio2023inversion}
M.~Delbracio and P.~Milanfar, ``Inversion by direct iteration: An alternative to denoising diffusion for image restoration,'' \emph{arXiv:2303.11435}, 2023.

\bibitem{choi2022perception}
J.~Choi, J.~Lee, C.~Shin, S.~Kim, H.~Kim, and S.~Yoon, ``Perception prioritized training of diffusion models,'' in \emph{CVPR}, 2022.

\bibitem{moser2023yoda}
B.~B. Moser, S.~Frolov, F.~Raue, S.~Palacio, and A.~Dengel, ``Yoda: You only diffuse areas. an area-masked diffusion approach for image super-resolution,'' \emph{arXiv:2308.07977}, 2023.

\bibitem{caron2021emerging}
M.~Caron, H.~Touvron, I.~Misra, H.~J{\'e}gou, J.~Mairal, P.~Bojanowski, and A.~Joulin, ``Emerging properties in self-supervised vision transformers,'' in \emph{CVPR}, 2021.

\bibitem{xia2023diffir}
B.~Xia, Y.~Zhang, S.~Wang, Y.~Wang, X.~Wu, Y.~Tian, W.~Yang, and L.~Van~Gool, ``Diffir: Efficient diffusion model for image restoration,'' \emph{arXiv:2303.09472}, 2023.

\bibitem{vaswani2017attention}
A.~Vaswani, N.~Shazeer, N.~Parmar, J.~Uszkoreit, L.~Jones, A.~N. Gomez, {\L}.~Kaiser, and I.~Polosukhin, ``Attention is all you need,'' \emph{NeurIPS}, vol.~30, 2017.

\bibitem{dosovitskiy2020image}
A.~Dosovitskiy, L.~Beyer, A.~Kolesnikov, D.~Weissenborn, X.~Zhai, T.~Unterthiner, M.~Dehghani, M.~Minderer, G.~Heigold, S.~Gelly \emph{et~al.}, ``An image is worth 16x16 words: Transformers for image recognition at scale,'' \emph{arXiv:2010.11929}, 2020.

\bibitem{kawar2021snips}
B.~Kawar, G.~Vaksman, and M.~Elad, ``Snips: Solving noisy inverse problems stochastically,'' \emph{NeurIPS}, vol.~34, 2021.

\bibitem{kawar2022denoising}
B.~Kawar, M.~Elad, S.~Ermon, and J.~Song, ``Denoising diffusion restoration models,'' \emph{NeurIPS}, vol.~35, 2022.

\bibitem{chung2022diffusion}
H.~Chung, J.~Kim, M.~T. Mccann, M.~L. Klasky, and J.~C. Ye, ``Diffusion posterior sampling for general noisy inverse problems,'' \emph{arXiv:2209.14687}, 2022.

\bibitem{wang2022zero}
Y.~Wang, J.~Yu, and J.~Zhang, ``Zero-shot image restoration using denoising diffusion null-space model,'' \emph{arXiv:2212.00490}, 2022.

\bibitem{fei2023generative}
B.~Fei, Z.~Lyu, L.~Pan, J.~Zhang, W.~Yang, T.~Luo, B.~Zhang, and B.~Dai, ``Generative diffusion prior for unified image restoration and enhancement,'' in \emph{CVPR}, 2023.

\bibitem{shocher2018zero}
A.~Shocher, N.~Cohen, and M.~Irani, ``“zero-shot” super-resolution using deep internal learning,'' in \emph{CVPR}, 2018.

\bibitem{li2024omnissr}
R.~Li, X.~Sheng, W.~Li, and J.~Zhang, ``Omnissr: Zero-shot omnidirectional image super-resolution using stable diffusion model,'' \emph{arXiv:2404.10312}, 2024.

\bibitem{lugmayr2022repaint}
A.~Lugmayr, M.~Danelljan, A.~Romero, F.~Yu, R.~Timofte, and L.~Van~Gool, ``Repaint: Inpainting using denoising diffusion probabilistic models,'' in \emph{CVPR}, 2022.

\bibitem{chung2022ccdf}
H.~Chung, B.~Sim, and J.~C. Ye, ``Come-closer-diffuse-faster: Accelerating conditional diffusion models for inverse problems through stochastic contraction,'' in \emph{CVPR}, 2022.

\bibitem{schwab2019deep}
J.~Schwab, S.~Antholzer, and M.~Haltmeier, ``Deep null space learning for inverse problems: convergence analysis and rates,'' \emph{Inverse Problems}, vol.~35, no.~2, 2019.

\bibitem{wang2023gan}
Y.~Wang, Y.~Hu, J.~Yu, and J.~Zhang, ``Gan prior based null-space learning for consistent super-resolution,'' in \emph{AAAI}, vol.~37, no.~3, 2023.

\bibitem{chung2022improving}
H.~Chung, B.~Sim, D.~Ryu, and J.~C. Ye, ``Improving diffusion models for inverse problems using manifold constraints,'' \emph{NeurIPS}, vol.~35, 2022.

\bibitem{song2022pseudoinverse}
J.~Song, A.~Vahdat, M.~Mardani, and J.~Kautz, ``Pseudoinverse-guided diffusion models for inverse problems,'' in \emph{ICLR}, 2022.

\bibitem{lin2024adaptive}
J.~Lin, Y.~Wang, Z.~Tao, B.~Wang, Q.~Zhao, H.~Wang, X.~Tong, X.~Mai, Y.~Lin, W.~Song \emph{et~al.}, ``Adaptive multi-modal fusion of spatially variant kernel refinement with diffusion model for blind image super-resolution,'' \emph{arXiv}, 2024.

\bibitem{chung2022mr}
H.~Chung, E.~S. Lee, and J.~C. Ye, ``Mr image denoising and super-resolution using regularized reverse diffusion,'' \emph{IEEE Transactions on Medical Imaging}, vol.~42, no.~4, 2022.

\bibitem{mao2023discdiff}
Y.~Mao, L.~Jiang, X.~Chen, and C.~Li, ``Disc-diff: Disentangled conditional diffusion model for multi-contrast mri super-resolution,'' \emph{arXiv:2303.13933}, 2023.

\bibitem{li2024rethinking}
G.~Li, C.~Rao, J.~Mo, Z.~Zhang, W.~Xing, and L.~Zhao, ``Rethinking diffusion model for multi-contrast mri super-resolution,'' \emph{arXiv:2404.04785}, 2024.

\bibitem{yue2022difface}
Z.~Yue and C.~C. Loy, ``Difface: Blind face restoration with diffused error contraction,'' \emph{arXiv:2212.06512}, 2022.

\bibitem{wang2021towards}
X.~Wang, Y.~Li, H.~Zhang, and Y.~Shan, ``Towards real-world blind face restoration with generative facial prior,'' in \emph{CVPR}, 2021.

\bibitem{yang2021gan}
T.~Yang, P.~Ren, X.~Xie, and L.~Zhang, ``Gan prior embedded network for blind face restoration in the wild,'' in \emph{CVPR}, 2021.

\bibitem{qiu2023diffbfr}
X.~Qiu, C.~Han, Z.~Zhang, B.~Li, T.~Guo, and X.~Nie, ``Diffbfr: Bootstrapping diffusion model towards blind face restoration,'' \emph{arXiv:2305.04517}, 2023.

\bibitem{wang2023dr2}
Z.~Wang, Z.~Zhang, X.~Zhang, H.~Zheng, M.~Zhou, Y.~Zhang, and Y.~Wang, ``Dr2: Diffusion-based robust degradation remover for blind face restoration,'' in \emph{CVPR}, 2023.

\bibitem{wang2023atmospheric}
X.~Wang, S.~L{\'o}pez-Tapia, and A.~K. Katsaggelos, ``Atmospheric turbulence correction via variational deep diffusion,'' in \emph{2023 IEEE 6th International Conference on MIPR}.\hskip 1em plus 0.5em minus 0.4em\relax IEEE, 2023.

\bibitem{nair2023ddpm}
N.~G. Nair, K.~Mei, and V.~M. Patel, ``At-ddpm: Restoring faces degraded by atmospheric turbulence using denoising diffusion probabilistic models,'' in \emph{WACV}, 2023.

\bibitem{xiao2023ediffsr}
Y.~Xiao, Q.~Yuan, K.~Jiang, J.~He, X.~Jin, and L.~Zhang, ``Ediffsr: An efficient diffusion probabilistic model for remote sensing image super-resolution,'' \emph{IEEE Transactions on Geoscience and Remote Sensing}, 2023.

\bibitem{rs14194834}
J.~Liu, Z.~Yuan, Z.~Pan, Y.~Fu, L.~Liu, and B.~Lu, ``Diffusion model with detail complement for super-resolution of remote sensing,'' \emph{Remote Sensing}, vol.~14, no.~19, 2022.

\bibitem{rs15092346}
A.~M. Ali, B.~Benjdira, A.~Koubaa, W.~Boulila, and W.~El-Shafai, ``Tesr: Two-stage approach for enhancement and super-resolution of remote sensing images,'' \emph{Remote Sensing}, vol.~15, no.~9, 2023.

\bibitem{xu2023dualdiffusion}
M.~Xu, J.~Ma, and Y.~Zhu, ``Dual-diffusion: Dual conditional denoising diffusion probabilistic models for blind super-resolution reconstruction in rsis,'' \emph{arXiv:2305.12170}, 2023.

\bibitem{khanna2023diffusionsat}
S.~Khanna, P.~Liu, L.~Zhou, C.~Meng, R.~Rombach, M.~Burke, D.~Lobell, and S.~Ermon, ``Diffusionsat: A generative foundation model for satellite imagery,'' \emph{ICLR}, 2024.

\bibitem{ganguli2022predictability}
D.~Ganguli, D.~Hernandez, L.~Lovitt, A.~Askell, Y.~Bai, A.~Chen, T.~Conerly, N.~Dassarma, D.~Drain, N.~Elhage \emph{et~al.}, ``Predictability and surprise in large generative models,'' in \emph{Proceedings of the 2022 ACM Conference on Fairness, Accountability, and Transparency}, 2022.

\bibitem{chen2020wavegrad}
N.~Chen, Y.~Zhang, H.~Zen, R.~J. Weiss, M.~Norouzi, and W.~Chan, ``Wavegrad: Estimating gradients for waveform generation,'' \emph{arXiv:2009.00713}, 2020.

\bibitem{cheng2023sampler}
Z.~Cheng, ``Sampler scheduler for diffusion models,'' \emph{arXiv:2311.06845}, 2023.

\bibitem{xiao2021tackling}
Z.~Xiao, K.~Kreis, and A.~Vahdat, ``Tackling the generative learning trilemma with denoising diffusion gans,'' \emph{arXiv:2112.07804}, 2021.

\bibitem{chen2023importance}
T.~Chen, ``On the importance of noise scheduling for diffusion models,'' \emph{arXiv:2301.10972}, 2023.

\bibitem{zhang2018unreasonable}
R.~Zhang, P.~Isola, A.~A. Efros, E.~Shechtman, and O.~Wang, ``The unreasonable effectiveness of deep features as a perceptual metric,'' in \emph{CVPR}, 2018.

\bibitem{liu2017rankiqa}
X.~Liu, J.~Van De~Weijer, and A.~D. Bagdanov, ``Rankiqa: Learning from rankings for no-reference image quality assessment,'' in \emph{ICCV}, 2017.

\bibitem{ma2017dipiq}
K.~Ma, W.~Liu, T.~Liu, Z.~Wang, and D.~Tao, ``dipiq: Blind image quality assessment by learning-to-rank discriminable image pairs,'' \emph{IEEE TIP}, vol.~26, no.~8, 2017.

\bibitem{yun2019cutmix}
S.~Yun, D.~Han, S.~J. Oh, S.~Chun, J.~Choe, and Y.~Yoo, ``Cutmix: Regularization strategy to train strong classifiers with localizable features,'' in \emph{CVPR}, 2019.

\bibitem{jing2022subspace}
B.~Jing, G.~Corso, R.~Berlinghieri, and T.~Jaakkola, ``Subspace diffusion generative models,'' in \emph{ECCV}.\hskip 1em plus 0.5em minus 0.4em\relax Springer, 2022.

\bibitem{kwon2023diffusion}
M.~Kwon, J.~Jeong, and Y.~Uh, ``Diffusion models already have a semantic latent space,'' in \emph{ICLR}, 2023.

\bibitem{wu2023uncovering}
Q.~Wu, Y.~Liu, H.~Zhao, A.~Kale, T.~Bui, T.~Yu, Z.~Lin, Y.~Zhang, and S.~Chang, ``Uncovering the disentanglement capability in text-to-image diffusion models,'' in \emph{CVPR}, 2023.

\bibitem{kim2022diffusionclip}
G.~Kim, T.~Kwon, and J.~C. Ye, ``Diffusionclip: Text-guided diffusion models for robust image manipulation,'' in \emph{CVPR}, 2022.

\end{thebibliography}

\vspace{-10.0 mm}

\begin{IEEEbiography}[{\includegraphics[width=1in,height=1in,clip,keepaspectratio]{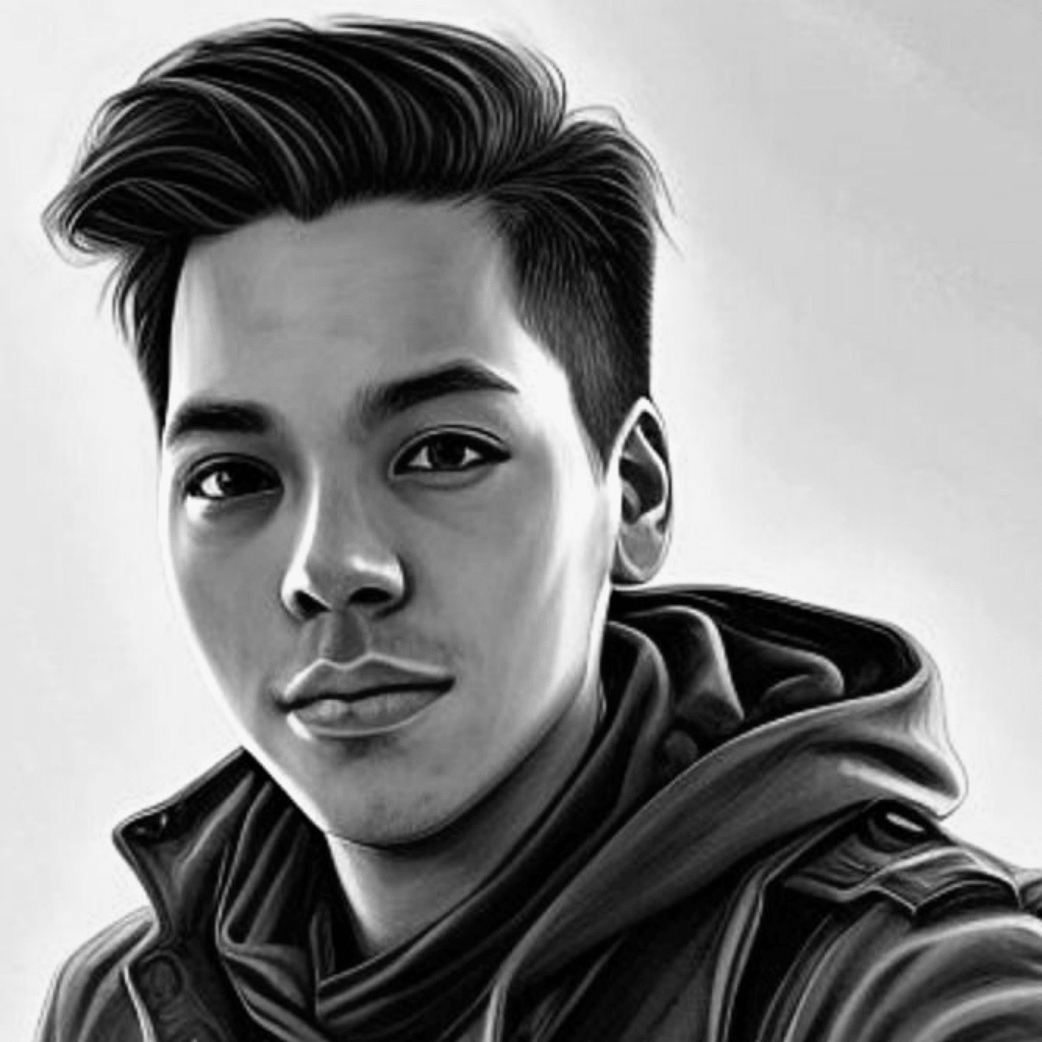}}]{Brian B. Moser}
 is a Ph.D. student at the TU Kaiserslautern and a research assistant at the German Research Center for Artificial Intelligence (DFKI) in Kaiserslautern. He received the M.Sc. degree in computer science from the TU Kaiserslautern in 2021. His research interests include image super-resolution and deep learning.
\end{IEEEbiography}

\vskip -4.1\baselineskip plus -1fil

\begin{IEEEbiography}[{\includegraphics[width=1in,height=1in,clip,keepaspectratio]{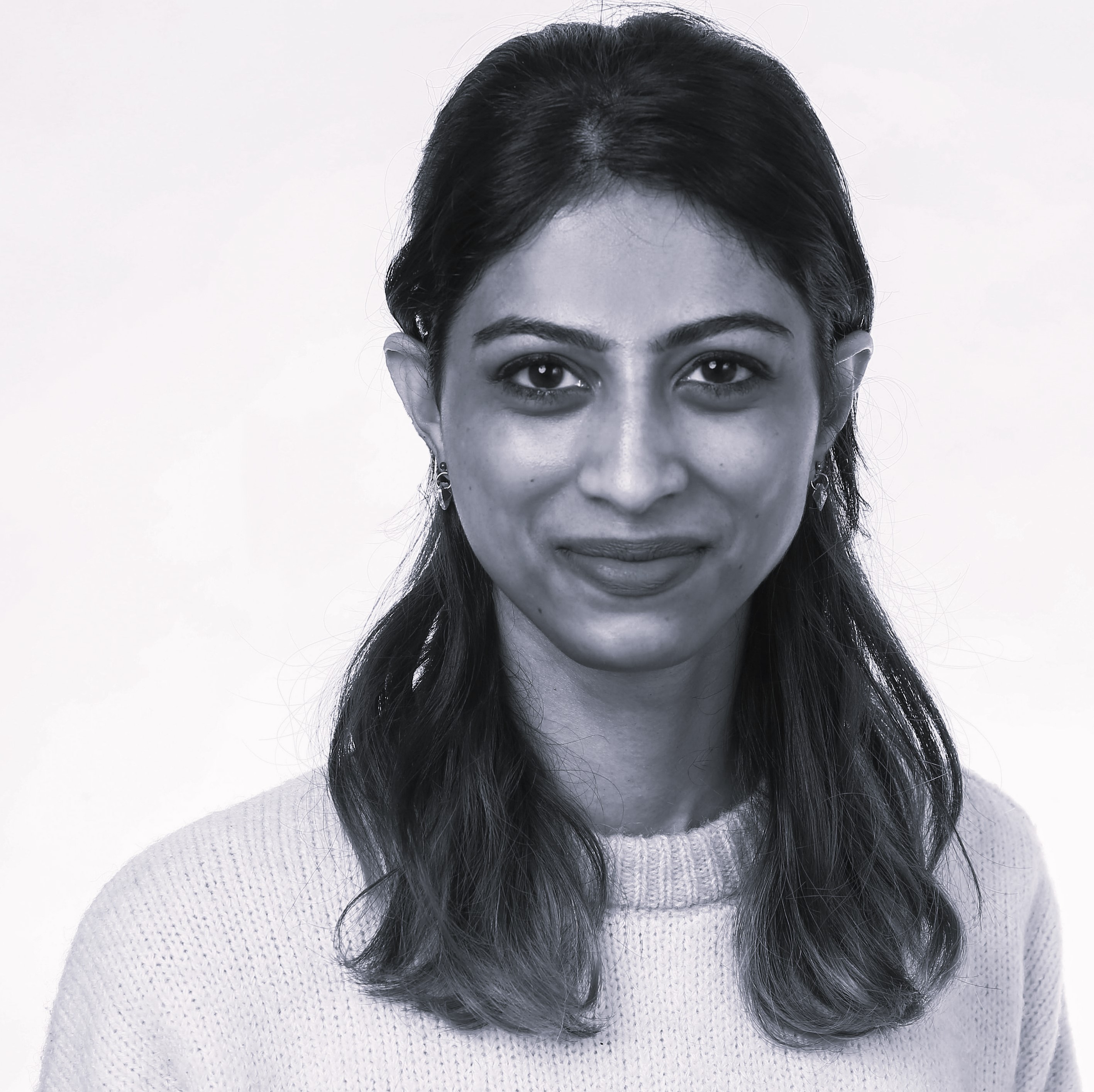}}]{Arundhati S. Shanbhag}
is a Master's student at the TU Kaiserslautern and research assistant at the German Research Center for Artificial Intelligence (DFKI) in Kaiserslautern. Her research interests include computer vision and deep learning. 
\end{IEEEbiography}

\vskip -4.1\baselineskip plus -1fil

\begin{IEEEbiography}[{\includegraphics[width=1in,height=1in,clip,keepaspectratio]{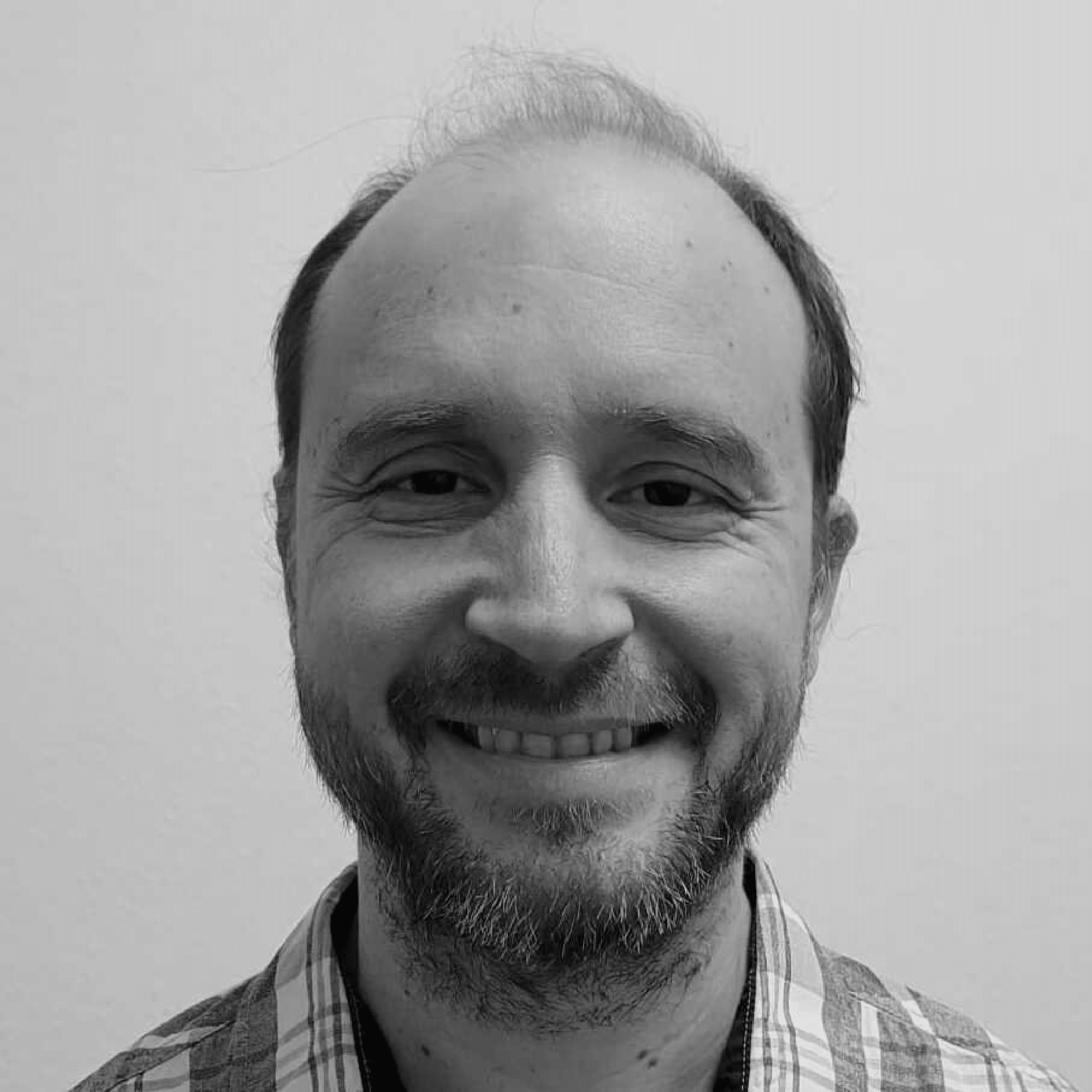}}]{Federico Raue}
 is a Senior Researcher at the German Research Center for Artificial Intelligence (DFKI) in Kaiserslautern. He received his Ph.D. at TU Kaiserslautern in 2018 and his M.Sc. in Artificial Intelligence from Katholieke Universiteit Leuven in 2005. His research interests include meta-learning and multimodal machine learning.
\end{IEEEbiography}

\vskip -4.1\baselineskip plus -1fil

\begin{IEEEbiography}[{\includegraphics[width=1in,height=1in,clip,keepaspectratio]{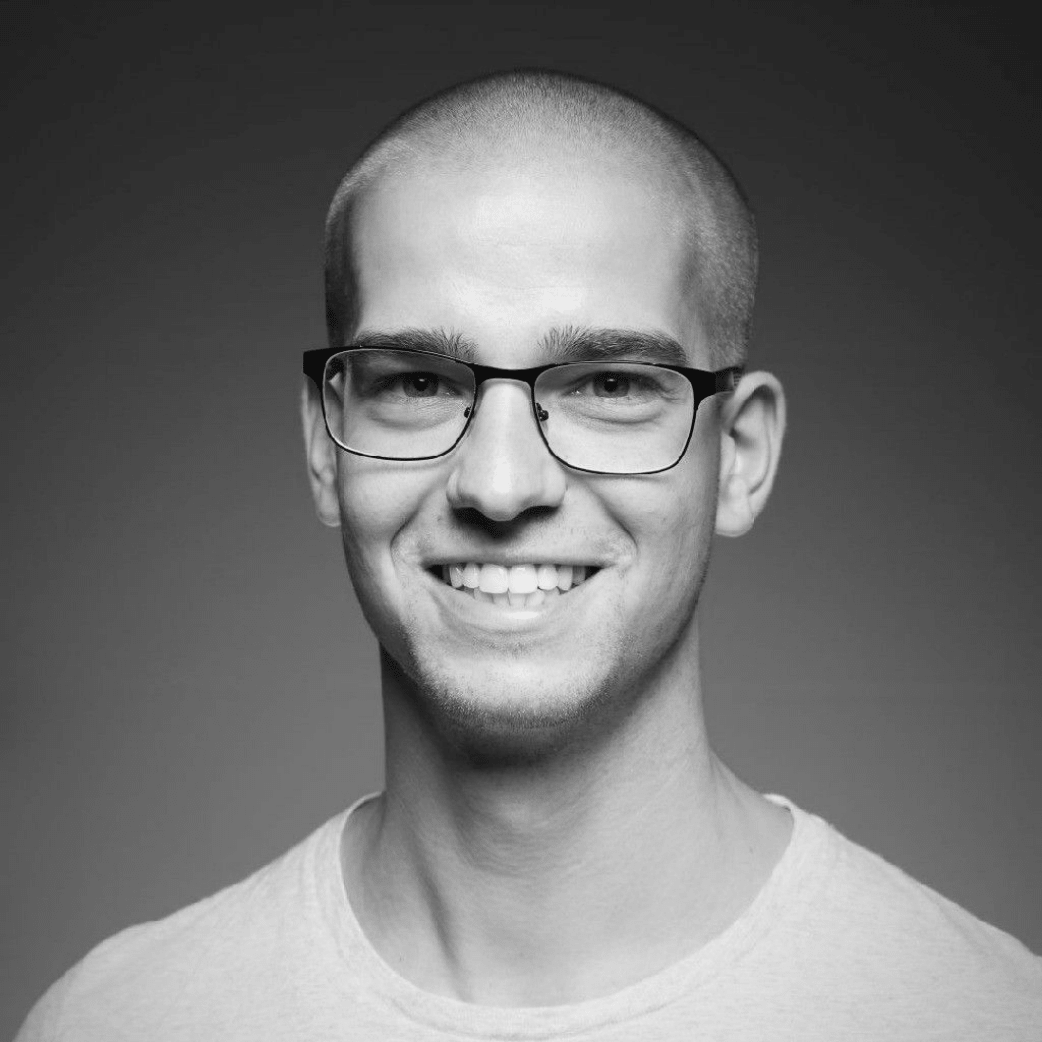}}]{Stanislav Frolov}
is a Ph.D. student at the TU Kaiserslautern and a research assistant at the German Research Center for Artificial Intelligence (DFKI) in Kaiserslautern. He received the M.Sc. degree in electrical engineering from the Karlsruhe Institute of Technology in 2017. His research interests include generative models and deep learning.
\end{IEEEbiography}

\vskip -4.1\baselineskip plus -1fil

\begin{IEEEbiography}[{\includegraphics[width=1in,height=1in,clip,keepaspectratio]{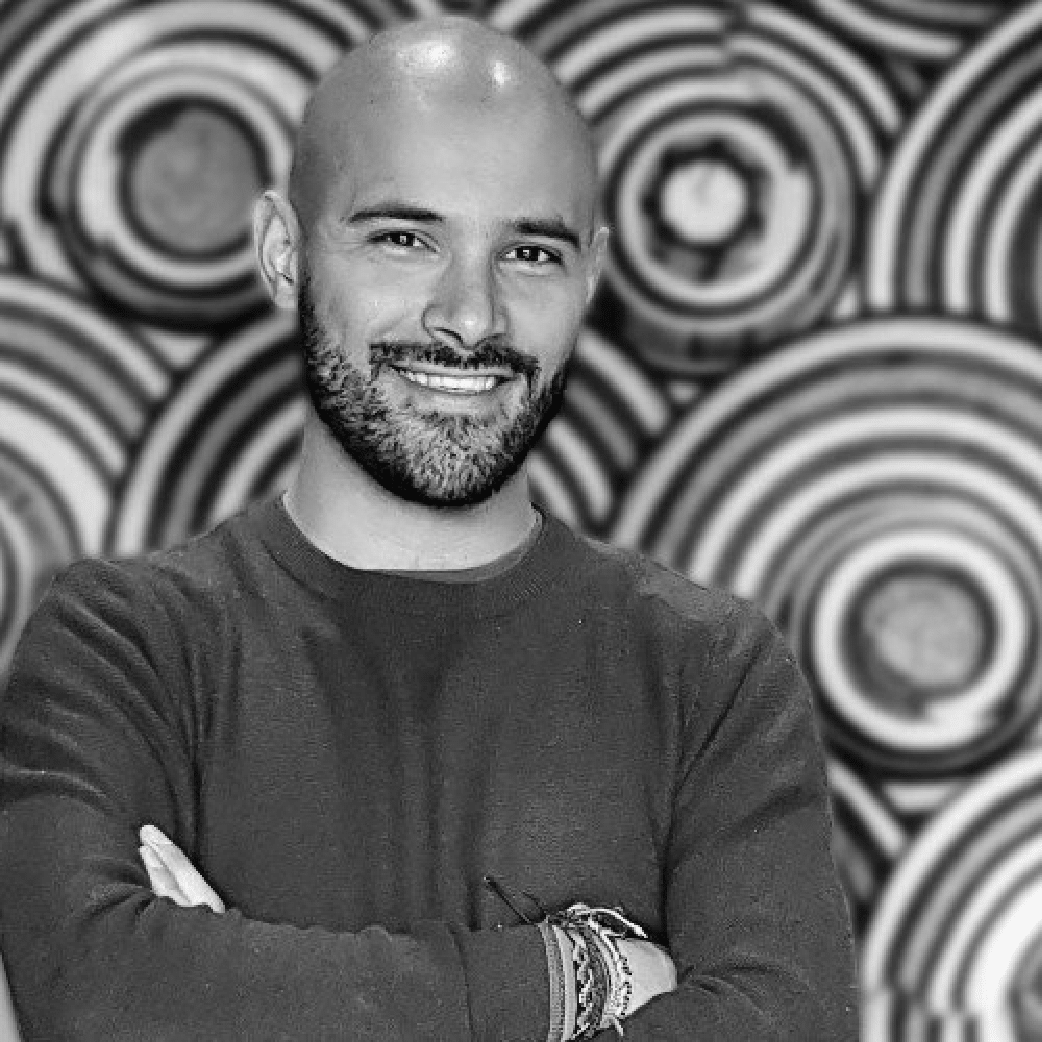}}]{Sebastian Palacio}
is a researcher in machine learning and head of the multimedia analysis and data mining group at the German Research Center for Artificial Intelligence (DFKI). His Ph.D. topic was about explainable AI with applications in computer vision. Other research interests include adversarial attacks, multi-task, curriculum, and self-supervised learning.
\end{IEEEbiography}

\vskip -4.1\baselineskip plus -1fil

\begin{IEEEbiography}[{\includegraphics[width=1in,height=1in,clip,keepaspectratio]{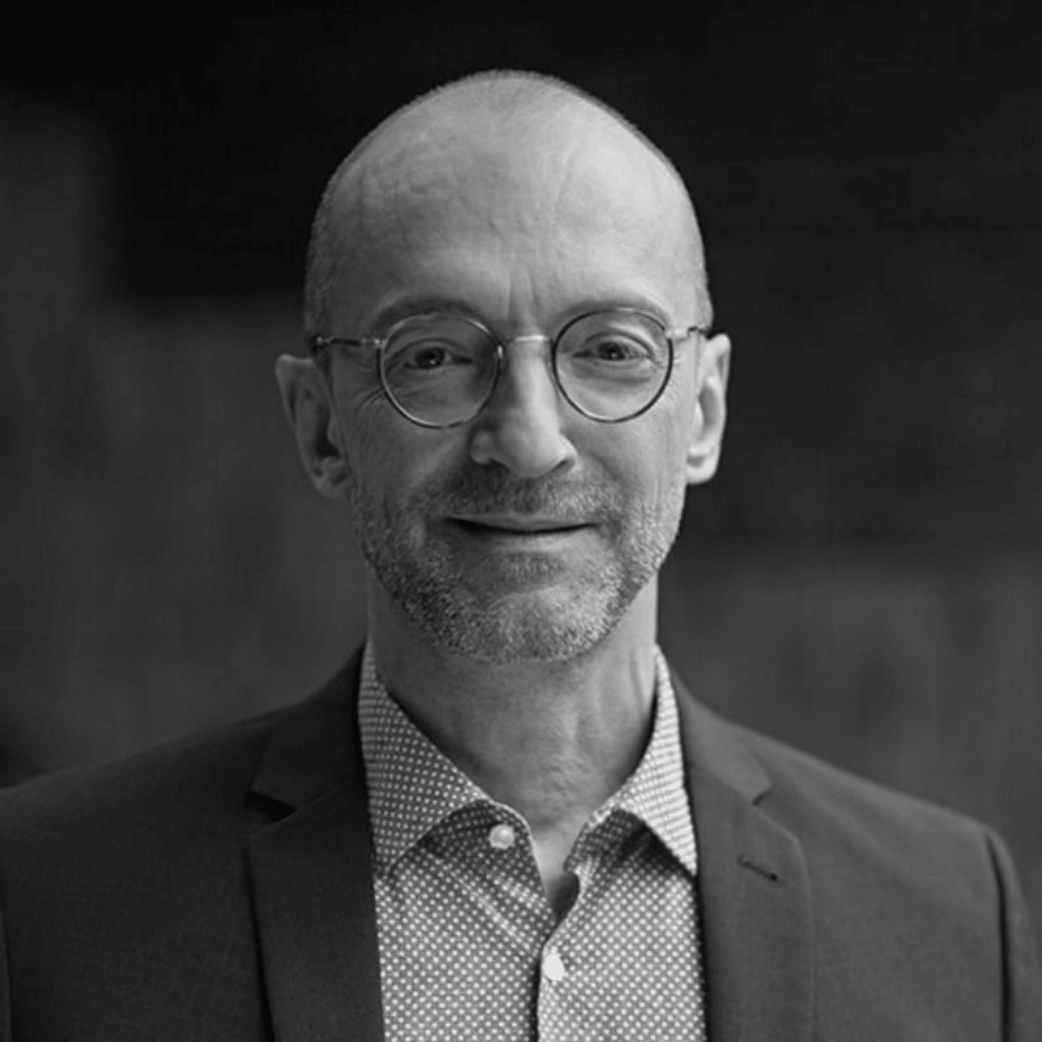}}]{Andreas Dengel}
is a Professor at the Department of Computer Science at TU Kaiserslautern and Executive Director of the German Research Center for Artificial Intelligence (DFKI) in Kaiserslautern, Head of the Smart Data and Knowledge  Services research area at DFKI and of the DFKI Deep Learning Competence Center. His research focuses on machine learning, pattern recognition, quantified learning, data mining, semantic technologies, and document analysis.
\end{IEEEbiography}

\vfill

\end{document}